%% file: elsarticle-template-harv.tex
\documentclass[preprint,12pt,authoryear]{elsarticle}

\usepackage{amssymb}
\usepackage{amsmath}

\journal{Preprint}
\usepackage{amsmath,amssymb}
\usepackage{algorithm}
\usepackage{algpseudocode}
\usepackage{hyperref}
\usepackage{xurl}
\usepackage{amsfonts}
\usepackage{booktabs}
\usepackage{tabularx}
\usepackage{array}
\usepackage{threeparttable}
\usepackage{multirow}
\usepackage{graphicx}
\usepackage{color}
\usepackage{enumitem}

\begin{document}

\begin{frontmatter}




\title{Bridging Transferability and Discriminability via Positive-Unlabeled Domain Adaptation for Building Damage Mapping Across Disaster Events}

\author[1]{Qirui Wang$^{\dagger}$}
\author[1]{Hengwei Zhao$^{\dagger}$}
\author[2]{Xiaoshu Yang}
\author[3]{Jingtao Li}
\author[4]{Bo Li}
\author[1]{Sorin Popescu}
\author[1]{Rusty Feagin}
\author[1]{Wenzhe Jiao\corref{cor1}}
\ead{wenzhe.jiao@ag.tamu.edu}

\affiliation[1]{
    organization={Department of Ecology and Conservation Biology,
                  Texas A\&M University},
    city={College Station},
    state={Texas},
    country={USA}
}

\affiliation[2]{
    organization={Department of Computer Science,
                  Rice University},
    city={Houston},
    state={Texas},
    country={USA}
}

\affiliation[3]{
    organization={Department of Computer Science and Technology,
                  University of Cambridge},
    city={Cambridge},
    state={England},
    country={UK}
}

\affiliation[4]{
    organization={Department of Civil \& Environmental Engineering,
                  Texas A\&M University},
    city={College Station},
    state={Texas},
    country={USA}
}

\cortext[cor1]{Corresponding author.}

\begin{abstract}
Mapping the damage to buildings following natural and human-caused disasters is essential for disaster response and humanitarian assistance. However, differences in hazard types, geographic environments, building characteristics, and imaging conditions cause cross-event distribution shifts that limit the generalization of deep learning models to new disaster events, even for geospatial foundation models (GFMs) pretrained on extensive Earth observation data. Unsupervised domain adaptation (UDA) can mitigate these shifts, but adaptation remains challenging when damage-proportion and damage-appearance shifts coexist and interact. We identify enhancing target-domain discriminability as key to addressing this challenge. To this end, inspired by positive-unlabeled (PU) learning, we introduce positive-unlabeled domain adaptation, a paradigm that incorporates PU weak supervision to enhance target-domain discriminability, requiring only a few labeled damaged buildings in the target domain. Building upon this straightforward intuition, we develop Positive-ANchored Domain Adaptation (PANDA), comprising two complementary and mutually reinforcing components: Pairwise Alignment with Intra-class Robustness (PAIR) for discriminative representation learning and Refine Gate and Verifier (RAV) for supervision refinement. On a global-scale multi-disaster benchmark, PANDA generally outperforms representative domain adaptation and PU learning methods. Results on the 2025 Los Angeles wildfires show its effectiveness in new disaster scenarios with few labeled damaged buildings. Further experiments demonstrate the potential of integrating PANDA with GFMs for cross-event adaptation. These findings highlight PANDA’s practical potential for rapid, reliable, and cross-event building damage mapping.
\end{abstract}



\end{frontmatter}

{\let\thefootnote\relax
\footnotetext{$^{\dagger}$\,Qirui Wang and Hengwei Zhao contributed equally to this work.}}

\section{Introduction}
\label{sec1}

Sudden-onset disasters can result in substantial human and socioeconomic losses~\citep{kucharczyk2021remote,elgharbawi2021damage}. Timely and reliable building damage assessments are essential for guiding emergency response and humanitarian assistance, while also supporting progress toward Sustainable Development Goal 11 (Sustainable Cities and Communities). Recent advances in high-resolution remote sensing imagery and deep learning have opened new opportunities for rapid, safe, accurate, and large-scale building damage mapping~\citep{gupta2019xbd,zheng2021building,tacskin2021review}.

Model transferability is essential for reliable building damage mapping across previously unseen disaster events characterized by diverse hazard types, geographic environments, and imaging conditions~\citep{zheng2024towards}. This capability is particularly critical in disaster response, where the extensive spatial coverage of affected areas~\citep{adriano2021learning}, stringent time constraints for emergency operations~\citep{fiedrich2000optimized}, and scarcity of labeled data immediately following disasters~\citep{ge2023rapid} necessitate the rapid deployment of models trained on historical disaster events. However, most deep learning models are typically developed under the independent and identically distributed (i.i.d.) assumption, whereby training data from historical disaster events (source domain) and deployment data from a new disaster event (target domain) are assumed to follow the same underlying distribution~\citep{koh2021wilds,ben2010theory}. In real-world disaster scenarios, this assumption may not always hold, and distribution shifts between training and deployment environments can consequently limit model generalization and lead to performance degradation in previously unseen disaster events~\citep{lin2022transfer,bouchard2022transfer,zheng2024towards,zheng2025neural}. Although recent geospatial foundation models (GFMs) have demonstrated promising generalization capabilities~\citep{liu2024remoteclip,brown2025alphaearth,herzog2026olmoearth}, their transferability across heterogeneous disaster events remains challenging, as further evidenced by the experimental results presented in Section~\ref{sec5.4.2}. Therefore, effective transfer strategies remain necessary to achieve reliable building damage mapping across previously unseen disaster events.

To mitigate distribution shifts, unsupervised domain adaptation (UDA) has emerged as a promising approach for transferring knowledge from labeled source domains to unlabeled target domains. Most UDA methods aim to improve model transferability by explicitly or implicitly mitigating cross-domain discrepancies through strategies such as discrepancy-based alignment~\citep{long2015learning,long2017deep}, adversarial learning~\citep{ganin2015unsupervised,tzeng2017adversarial}, and self-training~\citep{zou2018unsupervised,liang2020we}. What's more, depending on the availability of source-domain data during adaptation, UDA methods can be broadly categorized into source-available UDA and source-free UDA~\citep{fang2024source}. While source-available UDA requires access to source-domain data during adaptation, source-free UDA eliminates this requirement by leveraging knowledge encoded in a pretrained source model~\citep{liang2020we,tang2024source,fang2024source}. In post-disaster building damage mapping, access to historical high-resolution remote sensing imagery may be constrained by data licensing and sharing restrictions~\citep{clark2017towards}, while retrieving, transmitting, and processing such imagery can incur substantial time and computational costs. These practical constraints make source-free UDA particularly attractive for rapid and resource-efficient building damage mapping. Nevertheless, the absence of source-domain data during adaptation introduces additional challenges for effective cross-domain knowledge transfer~\citep{fang2024source}.

\begin{figure}[!t]
    \centering
    \includegraphics[width=1\linewidth]{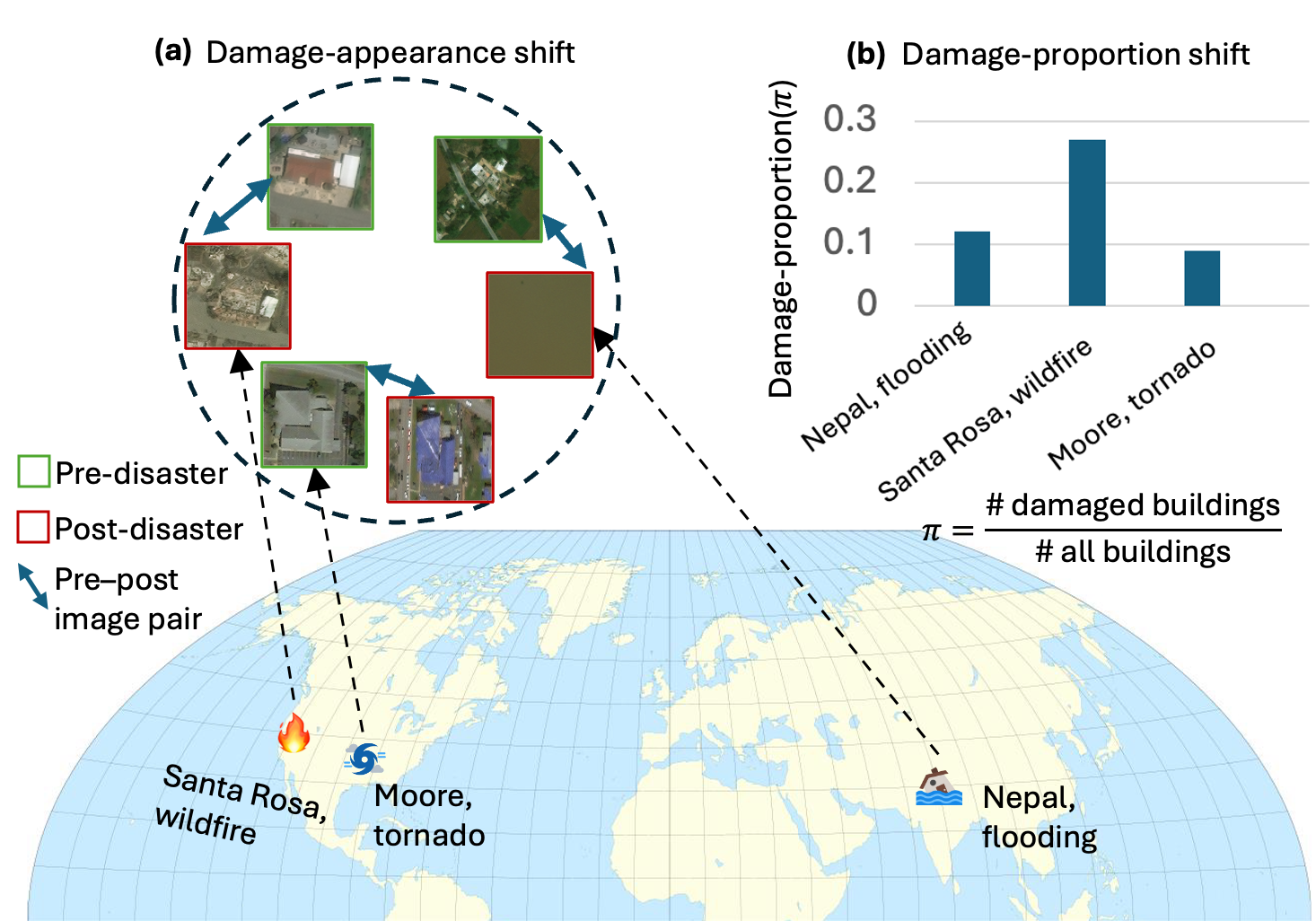}
    \caption{Coupled distribution shifts across disaster events in building damage mapping: (a) different building damage patterns across disaster events (damage-appearance shift), and (b) varying proportions of damaged buildings across events (damage-proportion shift).}
    \label{fig:shifts}
\end{figure}

Beyond these practical constraints, cross-event building damage mapping is further complicated by two interacting forms of cross-event distribution shift (Figure~\ref{fig:shifts}): (1) damage-proportion shift and (2) damage-appearance shift. To characterize these shifts, let $x$ denote the image input of a building instance, with subscripts $s$ and $t$ indicating the source and target domains, respectively. The corresponding data distributions can be expressed as mixtures of damaged and undamaged building instances:
\begin{equation}\label{eq:mix}
P_s(x) = \pi_s P_s^{d}(x) + (1-\pi_s) P_s^{n}(x), \qquad
P_t(x) = \pi_t P_t^{d}(x) + (1-\pi_t) P_t^{n}(x), 
\end{equation}
where the superscripts $d$ and $n$ denote the damaged and undamaged classes, respectively; $P^{d}$ and $P^{n}$ represent their corresponding class-conditional distributions; and $\pi_s$ and $\pi_t$ denote the proportions of damaged buildings in the source and target domains. Variations in disaster severity and the spatial distribution of affected buildings can lead to differences in damage proportions across disaster events (i.e., $\pi_s \neq \pi_t$), resulting in a damage-proportion shift. Meanwhile, variations in disaster types and building characteristics can alter the visual characteristics of damaged and undamaged buildings, leading to differences in class-conditional distributions (i.e., $P_s^{d}(x) \neq P_t^{d}(x)$ or $P_s^{n}(x) \neq P_t^{n}(x)$), referred to as damage-appearance shift. The coexistence of these shifts poses substantial challenges for UDA methods. For instance, aligning the overall source- and target-domain feature distributions (i.e., $P_s(z) \approx P_t(z)$, where $z$ denotes the learned representation) does not explicitly account for shifts in damage proportions or damage appearance distributions. In other words, such global alignment does not necessarily ensure that $P_s^d(z) \approx P_t^d(z)$ or that $\pi_s \approx \pi_t$. Consequently, although such alignment can enhance transferability, it does not necessarily preserve within-domain discriminability, defined as the ability of learned representations to distinguish between damaged and undamaged buildings. In particular, marginal alignment may inadvertently bring representations of different classes closer together~\citep{zhao2019learning} or suppress discriminative features essential for target-domain predictions~\citep{chen2019transferability}. Although source-domain supervision can promote source-domain discriminability, it does not necessarily ensure target-domain discriminability in the absence of target-domain supervision~\citep{qiang2025transferability}. Therefore, reliable building damage mapping across disaster events requires learning representations that jointly achieve cross-domain transferability and target-domain discriminability, as further supported by the information-theoretic analysis in Section~\ref{sec4}.

To enhance target-domain discriminability, this study seeks to leverage relatively accessible weak supervisory information through positive-unlabeled (PU) learning~\citep{10377190}, transforming it into reliable supervisory signals. In real-world disaster response, a small number of damaged buildings (P) can often be rapidly annotated, while the remaining buildings (U) are left unlabeled, naturally forming positive and unlabeled data. This setting is consistent with the basic formulation of PU learning, which aims to learn a binary classifier from P and U data. Existing PU learning approaches include two-step methods that identify reliable negative samples before classifier training~\citep{gong2018margin} and one-step methods based on risk estimation~\citep{zhao2022mapping,10174705}, density-ratio estimation~\citep{kato2019learning}, variational principles~\citep{10377190}, cost-sensitive learning~\citep{chen2021cost}, and label disambiguation~\citep{zhao2026noisy,zhang2019positive}. However, these methods primarily focus on learning from positive and unlabeled samples without explicitly considering how to effectively leverage knowledge acquired from historical disaster events.

\begin{figure}
    \centering
    \includegraphics[width=1\linewidth]{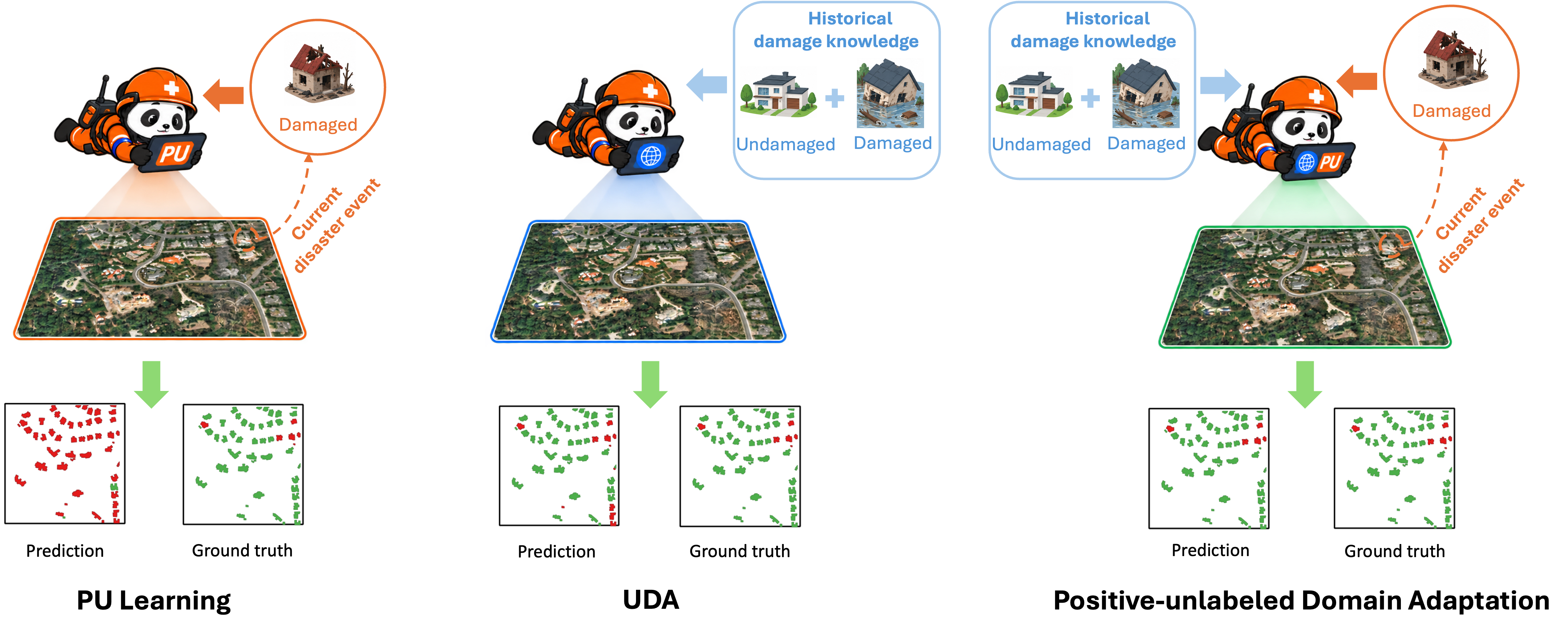}
    \caption{Comparison of PU learning, unsupervised domain adaptation and positive-unlabeled domain adaptation. For each paradigm, the left result shows the predicted building damage map, while the right result shows the ground-truth building damage map. Green and red denote undamaged and damaged buildings, respectively. ChatGPT (OpenAI, GPT-6.1 Sol) was used to generate the panda illustrations and the icons depicting damaged and undamaged buildings.}
    \label{fig:pcda}
\end{figure}

To bridge this gap, this study introduces a positive-unlabeled domain adaptation paradigm (Figure~\ref{fig:pcda}) that integrates domain adaptation with PU learning to jointly enhance cross-domain transferability and target-domain discriminability. These two objectives are complementary, as focusing on either alone may be insufficient to address the coupled distribution shifts encountered in cross-event building damage mapping. Furthermore, this study demonstrates that learning discriminative representations during adaptation is critical for effectively reconciling transferability and discriminability. Building on this insight, we propose Positive-ANchored Domain Adaptation (PANDA), a unified source-free framework that jointly enhances cross-domain transferability and target-domain discriminability for cross-event building damage mapping. The main contributions of this study are summarized as follows:

\begin{enumerate}[label=(\arabic*)]
\item Two coupled distribution shifts in cross-event building damage mapping, namely damage-proportion and damage-appearance shifts, are identified, and the importance of jointly achieving cross-domain transferability and target-domain discriminability for reliable adaptation is highlighted.

\item PANDA is proposed to integrate source-free domain adaptation with positive-unlabeled supervision for transferring historical damage knowledge and enhancing target-domain discriminability. Specifically, Pairwise Alignment with Intra-class Robustness (PAIR) is introduced to learn discriminative representations through class-consistent alignment while mitigating the influence of unreliable sample pairs. Refine Gate and Verifier (RAV) is designed to leverage the representations learned by PAIR to progressively refine supervisory signals for unlabeled target samples, thereby enhancing target-domain discriminability.

\item Extensive experiments are conducted on benchmark datasets spanning diverse disaster events worldwide and local-scale case studies of the 2025 Los Angeles wildfires. The results demonstrate the effectiveness of PANDA in cross-event building damage mapping and highlight its potential for accurate and large-scale post-disaster damage assessment.
\end{enumerate}

The remainder of this paper is organized as follows. Section~\ref{sec2} describes the study area and data. Section~\ref{sec3} details the proposed method, including the PANDA framework and its key components. Section~\ref{sec4} analyzes the importance of target-domain discriminability for domain adaptation under coupled shifts from an information-theoretic perspective. Section~\ref{sec5} presents and discusses the experimental results. Section~\ref{sec6} concludes the paper.

\section{Study area and data}
\label{sec2}

\begin{figure}[!t]
    \centering
    \includegraphics[width=\linewidth]{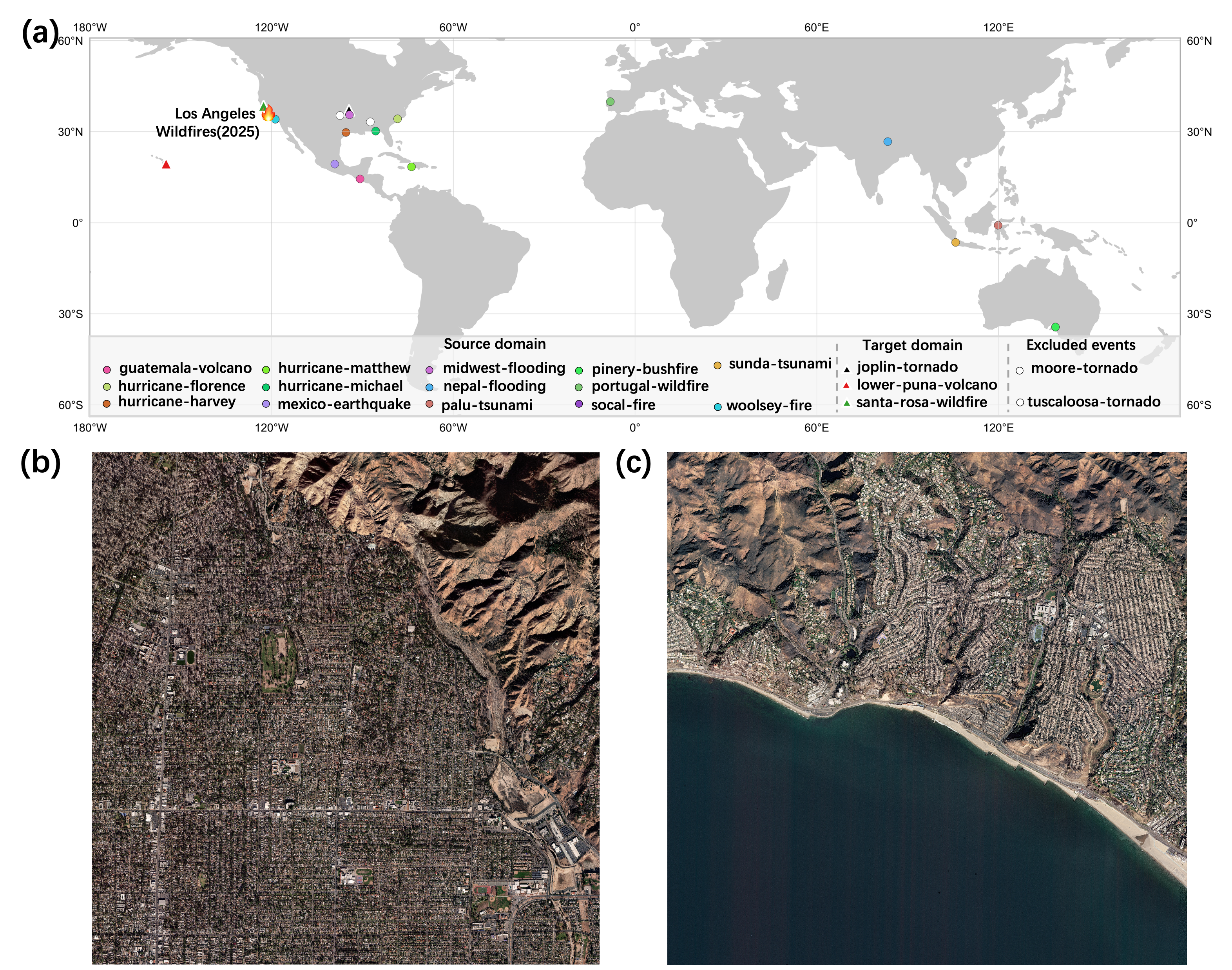}
    \caption{Benchmark and local study areas for cross-event building damage mapping: (a) spatial distribution of all disaster events considered in this study; (b) local study area of the 2025 Eaton Fire; and (c) local study area of the 2025 Palisades Fire.}
    \label{fig:studyarea}
\end{figure}

In this study, we evaluated PANDA from two complementary perspectives: a benchmark evaluation using datasets from diverse regions worldwide and two local-scale case studies. For the benchmark evaluation, we used the xView2 Building Damage Assessment (xBD) dataset~\citep{gupta2019xbd} as the benchmark for quantitative evaluation, with the spatial distribution of the considered disaster events shown in Figure~\ref{fig:studyarea}(a). For the local-scale studies, two representative affected areas within the January 2025 Los Angeles wildfires were selected as real-world case studies, as shown in Figure~\ref{fig:studyarea}(b) and Figure~\ref{fig:studyarea}(c). These local cases are geographically and temporally distinct from the global-scale study area, providing realistic scenarios for further validating the practical applicability of PANDA to previously unseen disaster events.

\subsection{Global benchmark study areas}
\label{sec2.1}

We used the xBD dataset~\citep{gupta2019xbd} for the benchmark evaluation. The xBD dataset contains 11,034 pairs of bitemporal high-resolution optical satellite images, covering a total area of 45,361.79 km² and including 850,736 building instances. The dataset includes 19 disaster events distributed across four continents, including Asia, Europe, North America, and Oceania (Figure~\ref{fig:studyarea}(a)). These events cover six disaster types: earthquake, wildfire, volcano, storm, flooding, and tsunami. The images were collected from multiple satellite platforms, such as WorldView-2 and WorldView-3, and therefore have various sub-meter spatial resolutions. Following previous studies on rapid building damage mapping~\citep{xia2021building}, the original annotations in xBD were converted into a binary classification setting, where major damage and destroyed were mapped to the damaged class, while minor damage and no damage were mapped to the undamaged class.

To evaluate cross-event adaptation, we reorganized the xBD dataset at the event level. Samples from the same disaster event can be approximately regarded as independently and identically distributed (i.i.d.), whereas samples from different events often differ in geographic location, disaster type, and event time. Event-level partitioning therefore enables a more appropriate evaluation of the model's cross-event adaptation capability from historical disasters to previously unseen disaster events. Among the 19 events in xBD, the Joplin tornado, Lower Puna volcano and Santa Rosa wildfire were used as target domains; the Moore and Tuscaloosa tornadoes were completely removed to ensure that the source domain contained no tornado-related samples; and the remaining 14 events were used as the source domain. This design covers two levels of cross-event adaptation difficulty: with all tornado events absent from source training, the Joplin case represents transfer to an entirely unseen disaster type, whereas the Lower Puna and Santa Rosa cases evaluate cross-event adaptation within disaster types already present in the source domain. Each target-domain event was divided into 80\% training and 20\% test subsets. Methods requiring target-domain positive samples used 400 labeled damaged buildings and all unlabeled buildings; other methods used the same training samples without any target-domain labels during adaptation.

\subsection{Local-scale study areas}
\label{sec2.2}

In January 2025, a series of wildfires swept across Los Angeles County in Southern California. Severe dry conditions, accumulated fuels, invasive grasses, and strong Santa Ana winds accelerated fire spread, consuming 55,082 acres and causing extensive damage to buildings and infrastructure. The fires claimed at least 29 lives and destroyed more than 16,251 structures. Among these events, the Eaton (Figure~\ref{fig:studyarea}(b)) and Palisades Fires (Figure~\ref{fig:studyarea}(c)) were among the most destructive, burning approximately 14,000 and 23,700 acres, respectively~\citep{li2025economic}, as shown in Figure~\ref{fig:studyarea}.

We collected satellite imagery over the affected areas of the Eaton and Palisades Fires from multiple data sources. The pre-disaster images were both obtained from Bing Maps. For the post-disaster images, WorldView-3 imagery acquired on January 14, 2025 was used for the Eaton Fire, with an original GSD of 0.32 m, while WorldView-2 imagery acquired on the same date was used for the Palisades Fire, with an original GSD of 0.57 m. All imagery was subsequently resampled to a uniform GSD of 0.3 m, with final image dimensions of 16,932 × 17,075 pixels for the Eaton Fire and 17,314 × 17,428 pixels for the Palisades Fire. The initial ground-truth labels were obtained from the Humanitarian Data Exchange datasets released by the Microsoft AI for Good Lab(\url{https://data.humdata.org/dataset/eaton-fire-altadena-damage-assessment-from-1-10}) and were subsequently reviewed and corrected by domain experts to improve their reliability for evaluation. For each local case, methods requiring target-domain positive samples used only 50 labeled damaged buildings (<1\% of all buildings in each event) and 2,000 unlabeled buildings; other methods used the same training samples without any target-domain labels during adaptation.

\section{Methodology}
\label{sec3}

\subsection{Problem reformulation for cross-event building damage mapping}
\label{sec3.1}

Building damage is inherently characterized at the instance level rather than the pixel level, with each building assigned a single damage status. Accordingly, building damage mapping can be naturally decomposed into two subtasks: (1) building localization to delineate individual building footprints and (2) building-level damage classification to determine the damage status of each identified building. Benefiting from well-established methods and publicly available large-scale building footprint datasets, building localization has become increasingly feasible across different regions~\citep{wang2024sampolybuild,li2024hd}. Consequently, the primary challenge in cross-event building damage mapping lies in achieving cross-event building-level damage classification, which is complicated by the coexistence of damage-proportion and damage-appearance shifts across disaster events.

\subsection{Building localization}
\label{sec3.2}

A flexible strategy is adopted to obtain reliable building footprints across disaster-affected regions: (1) existing building footprint products are used when reliable and up-to-date data are available for the affected region; and (2) otherwise, particularly in newly affected regions lacking reliable building footprint products, FootprintSAM, a building footprint extraction model built upon the Segment Anything Model (SAM)~\citep{kirillov2023segment}, is developed to automatically delineate building footprints from the corresponding pre-disaster imagery.

\begin{figure}[!t]
    \centering
    \includegraphics[width=1\linewidth]{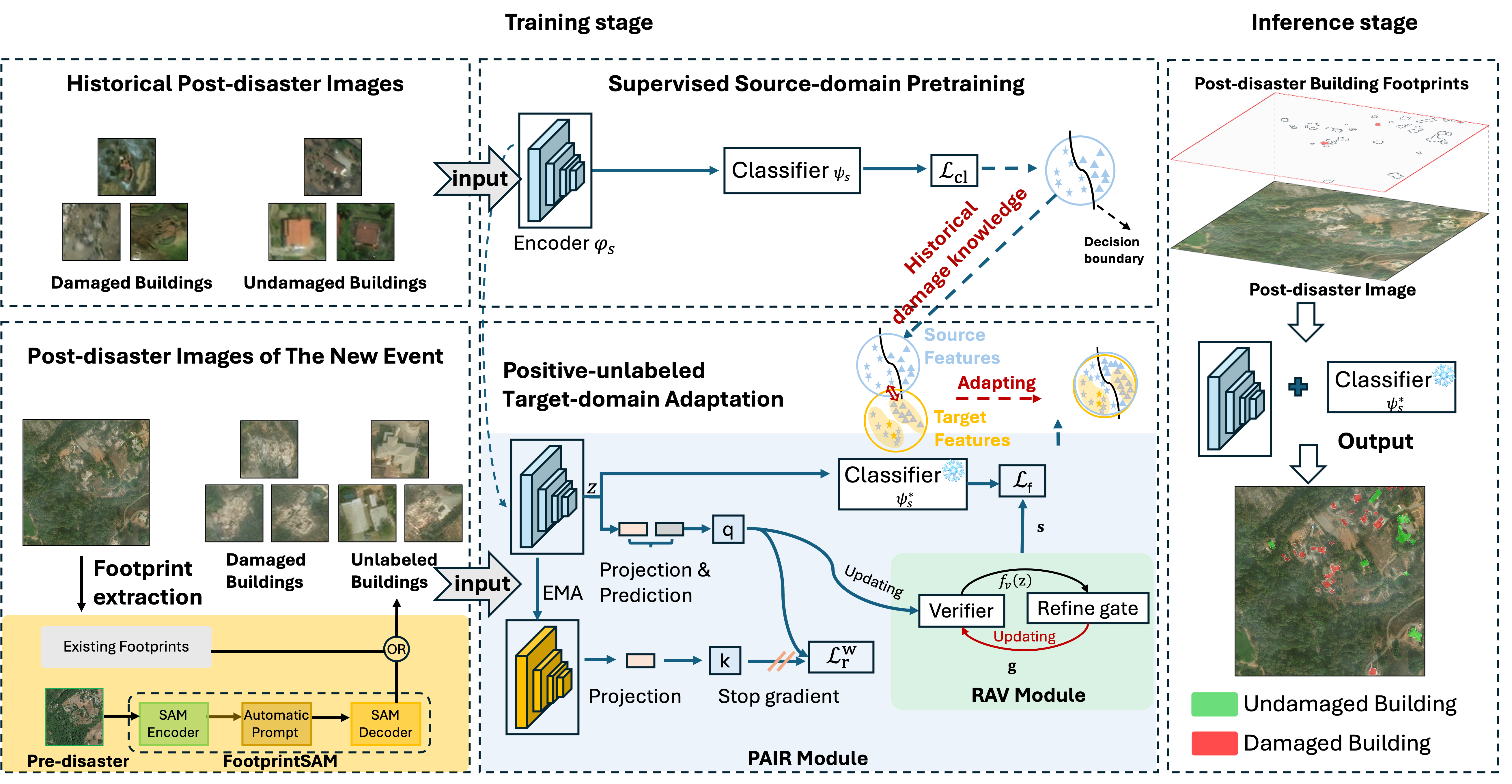}
    \caption{Overview of the proposed PANDA framework, consisting of supervised source-domain pretraining and positive-unlabeled target-domain adaptation, in which PAIR and RAV jointly achieve cross-domain adaptation.}
    \label{fig:panda}
\end{figure}

The overall architecture of FootprintSAM is illustrated in Figure~\ref{fig:panda}. As a vision foundation model, SAM exhibits strong cross-region generalization capabilities. However, its class-agnostic segmentation requires appropriate prompts to identify and delineate individual buildings. To address this limitation, inspired by SAMPolyBuild~\cite{wang2024sampolybuild}, FootprintSAM integrates MS-DETR~\citep{zhao2024ms} as an automatic prompt generation module that leverages cross-attention to aggregate contextual image features and generate building-specific prompts. FootprintSAM is trained offline using pre-disaster imagery from historical disaster events and the corresponding building footprint annotations, with the detection loss of the prompt generation module and the segmentation loss of SAM jointly optimized. By combining extensive historical training data with the strong generalization capabilities of SAM, FootprintSAM can be directly applied to pre-disaster imagery of newly affected regions without region-specific retraining or manually specified prompts. Furthermore, experimental results demonstrate that MS-DETR-based prompt generation achieves more accurate building footprint delineation than the convolutional region proposal network employed in SAMPolyBuild (Section~\ref{sec5.4.5}).

\subsection{Building-level damage classification}
\label{sec3.3}

To jointly enhance cross-domain transferability and target-domain discriminability for building damage mapping across disaster events, the proposed PANDA framework (Figure~\ref{fig:panda}) consists of two stages: (1) supervised source-domain pretraining to acquire historical damage knowledge and (2) positive-unlabeled target-domain adaptation to transfer this knowledge while enhancing target-domain discriminability. PANDA requires only a small number of labeled damaged-building samples from newly affected regions, without relying on target-domain negative labels or prior knowledge of the target-domain class distribution.

\subsubsection{Supervised source-domain pretraining}
\label{sec3.3.1}

Following the decoupling described in Section~\ref{sec3.1}, source-domain pretraining focuses on building-level damage classification using historical post-disaster imagery to acquire damage knowledge.

Specifically, let $\mathcal{D}_s=\{(x_i^{s},y_i^{s})\}_{i=1}^{N_s}$ denote the labeled historical disaster dataset, where $x_i^{s}$ is the post-disaster image patch of the $i$-th building, $y_i^{s}\in\{0,1\}^2$ denotes its one-hot-encoded damage label, with $[0,1]^\top$ indicating damaged and $[1,0]^\top$ indicating undamaged, and $N_s$ is the number of labeled source-domain samples. The model consists of a feature extractor $\varphi$ and a damage classifier $\psi_s$. We pretrain both components by minimizing the cross-entropy loss:
\begin{equation}
\mathcal{L}_{\mathrm{cl}}=-\frac{1}{N_s}\sum_{i=1}^{N_s}(y_i^s)^{\top}\log {\psi_s}\left(\varphi(x_i^s)\right).
\label{eq:source_pretraining}
\end{equation}
Following the source hypothesis freezing strategy~\citep{liang2020we}, the pretrained damage classifier, denoted by $\psi_s^*$, remains frozen throughout the subsequent positive-anchored domain adaptation, thereby preserving the building damage knowledge acquired from historical disaster events. This knowledge is subsequently transferred to newly affected regions without requiring access to historical source-domain data.

\subsubsection{Positive-unlabeled target-domain adaptation}
\label{sec3.3.2}

Building on the preserved historical damage knowledge, PANDA incorporates PU supervision to enhance target-domain discriminability using only a small number of labeled damaged buildings from the target domain. For simplicity, the target-domain superscript $t$ is omitted hereafter. Let $x_i$ denote the post-disaster image of the $i$-th target-domain building. To accommodate potentially unreliable PU supervisory signals from the target domain, a flexible supervision strategy is adopted, formulated as follows:

\begin{equation}
\operatorname{CE}\bigl(s_i, {\psi_s^*}\left(z_i\right)\bigr)=-{s}_i^{\top}\log {\psi_s^*}\left(z_i\right),
\label{eq:target_ce}
\end{equation}
where $z_i=\varphi(x_i)$ denotes the feature representation of $x_i$ extracted by the encoder $\varphi$. Here, $s_i \in [0,1]^2$ denotes the supervisory signal for $x_i$. For labeled damaged buildings in the positive set $\mathcal{P}_t$, $s_i$ is assigned the ground-truth positive label $\mathbf{p}=[0,1]^\top$. For buildings in the unlabeled set $\mathcal{U}_t$, $s_i$ denotes a pseudo-target that is progressively refined during adaptation. Reliable supervisory signals are essential for achieving target-domain discriminability during adaptation. However, identifying reliable supervisory signals, in turn, depends on the discriminability of target-domain representations. This interdependence poses a challenge to effectively leveraging target-domain PU supervision. To address this challenge, PANDA further optimizes the target-domain feature extractor through two complementary components: Pairwise Alignment with Intra-class Robustness (PAIR), which aims to learn discriminative representations, and Refine Gate and Verifier (RAV), which aims to obtain more accurate supervisory signals $s_i$. The discriminative representations learned by PAIR facilitate the refinement of pseudo-targets for unlabeled buildings in RAV, while the refined supervisory signals provide more reliable class assignments for sample pairing in PAIR.

\begin{algorithm}[!t]
\caption{Pairwise Alignment with Intra-class Robustness (PAIR)}
\label{alg:pair}
\begin{algorithmic}[1]
\Require Positive and unlabeled mini-batches $\mathcal{B}_P$ and $\mathcal{B}_U$, feature extractor $\varphi$, online and target branches $\mathcal{F}$ and $\mathcal{F}'$, and frozen classifier $\psi_s^{*}$

\Statex \textit{// Building representations and class assignments}
\For{$x_i \in \mathcal{B}_P \cup \mathcal{B}_U$}
    \State $\tilde{y}_i \gets \psi_s^{*}(\varphi(x_i))$ \Comment{Data augmentation is omitted for simplicity}
    \Statex \hspace{\algorithmicindent}\textit{// Online query and target key}
    \State $q_i \gets \mathcal{F}(x_i)$, \quad $k_i \gets \mathcal{F}'(x_i)$ \Comment{Data augmentation is omitted}
    \State $\tilde{q}_i \gets q_i/\|q_i\|_2$,
           \quad $\tilde{k}_i \gets k_i/\|k_i\|_2$
\EndFor

\Statex \textit{// Robust pairwise alignment}
\State Compute $\overline{\mathcal{L}}_{\mathrm{r}}$ over sampled pairs
       using Eq.~\eqref{eq:pair_loss}.
\State \Return $\{z_i,\tilde{q}_i\}$ and $\overline{\mathcal{L}}_{\mathrm{r}}$
\end{algorithmic}
\end{algorithm}

\textbf{Pairwise Alignment with Intra-class Robustness (PAIR).}
PAIR is proposed to learn discriminative target-domain representations during cross-event adaptation (Algorithm~\ref{alg:pair}). It integrates a non-contrastive representation learning framework BYOL~\citep{grill2020bootstrap} for domain adaptation with a loss function designed to mitigate the influence of noise arising from weak PU supervision. Following the source hypothesis freezing strategy, PAIR keeps the source-domain classifier $\psi_s^*$ fixed during adaptation to facilitate the transfer of historical damage knowledge while updating the feature extractor $\varphi$ to learn discriminative target-domain representations. Given a target-domain building image $x_i$, two randomly augmented views, $v_i$ and $v'_i$, are generated. The online branch $\mathcal{F}$ processes $v_i$ through an encoder, a projection head, and a prediction head to produce the query representation $q_i$. The target branch $\mathcal{F}'$ processes $v'_i$ through its encoder and projection head to produce the key representation $k_i$. The target branch is excluded from backpropagation, and its parameters are updated by an exponential moving average of the corresponding online parameters with momentum coefficient $m$. The $\ell_2$-normalized query and key representations are given by $\tilde{q}_i = q_i / \|q_i\|_2$ and $\tilde{k}_i = k_i / \|k_i\|_2$, respectively.

Three types of loss functions can be employed within the PANDA framework to learn discriminative representations: a self-supervised loss $\mathcal{L}_{\mathrm{r}}^{\mathrm{s}}$, a fully supervised loss $\mathcal{L}_{\mathrm{r}}^{\mathrm{f}}$, and a weakly supervised loss $\mathcal{L}_{\mathrm{r}}^{\mathrm{w}}$. $\mathcal{L}_{\mathrm{r}}^{\mathrm{s}}$ is the representation learning loss adopted in the BYOL framework, which can be formulated as:
\begin{equation}
\mathcal{L}_{\mathrm{r}}^{\mathrm{s}}(x_i)=2\left(1-\langle\tilde{{q}}_i,\tilde{{k}}_i\rangle\right).
\end{equation}
However, as a self-supervised loss, $\mathcal{L}_{\mathrm{r}}^{\mathrm{s}}$ cannot fully exploit the PU supervisory information available from target domain. $\mathcal{L}_{\mathrm{r}}^{\mathrm{f}}$ leverages reliable supervisory information as follows:
\begin{equation}
\mathcal{L}_{\mathrm{r}}^{\mathrm{f}}(x_i,x_j)=2\left(1-\langle\tilde{{q}}_i,\tilde{{k}}_j\rangle\right)\mathbb{I}(y_i=y_j),
\end{equation}
where $y_i$ and $y_j$ denote the class labels of samples $x_i$ and $x_j$, respectively. However, in the absence of accurate labels, replacing ground-truth supervision with imperfect PU supervision inevitably introduces supervisory noise, which can adversely affect the domain adaptation process. To mitigate the adverse effects of potentially unreliable PU supervision, PAIR employs the weakly supervised loss $\mathcal{L}_{\mathrm{r}}^{\mathrm{w}}$~\citep{zhao2026noisy} to align representations of samples predicted to belong to the same class:
\begin{equation}
\mathcal{L}_{\mathrm{r}}^{\mathrm{w}}(x_i,x_j)=2\sqrt{1-\langle\tilde{q}_i,\tilde{k}_j\rangle}\mathbb{I}(\tilde{y}_i=\tilde{y}_j),
\label{eq:pair_loss}
\end{equation}
where $\tilde{y}_i$ and $\tilde{y}_j$ denote the class labels predicted by $\psi_s^*$ for $x_i$ and $x_j$, respectively. Suppose that $x_j$ forms a reliable pair with $x_i$ with higher representation similarity, whereas $x_m$ forms an unreliable pair with $x_i$ with lower representation similarity. A gradient-based analysis further reveals that
\begin{equation}
\left\|
\frac{\partial\mathcal{L}_{\mathrm{r}}^{\mathrm{w}}(x_i,x_m)}{\partial q_i}
\right\|_2^2
=
\frac{1+\langle\tilde{q}_i,\tilde{k}_m\rangle}{\|q_i\|_2^2}
<
\frac{1+\langle\tilde{q}_i,\tilde{k}_j\rangle}{\|q_i\|_2^2}
=
\left\|
\frac{\partial\mathcal{L}_{\mathrm{r}}^{\mathrm{w}}(x_i,x_j)}{\partial q_i}
\right\|_2^2.
\label{eq:pair_gradient}
\end{equation}
In other words, unreliable sample pairs produce smaller gradient norms than reliable ones, enabling PAIR to achieve robust intra-class representation alignment despite the presence of unreliable supervisory signals.

\begin{algorithm}[!t]
\caption{Refine Gate and Verifier (RAV)}
\label{alg:rav}
\begin{algorithmic}[1]
\Require Positive and unlabeled mini-batches $\mathcal{B}_P$ and
         $\mathcal{B}_U$, PAIR representations $\{z_i,\tilde{q}_i\}$,
         Verifier $f_v$, class prototypes
         $\{\boldsymbol{\mu}_c\}$,
         the initialized $\{h_i\}$, momentum $\beta$
\State Update the class-specific thresholds $\tau=[\tau_0,\tau_1]$ using SAT after each mini-batch
\For{$x_i \in \mathcal{B}_P \cup \mathcal{B}_U$}
    \Statex \textit{// Prototype update}
    \State $\boldsymbol{\mu}_c \gets \alpha\boldsymbol{\mu}_c + (1-\alpha)\tilde{q}_i$
    \Statex \textit{// Refine Gate for} $g_i$ \textit{and} $s_i$
    \State $r_i \gets \operatorname{OneHot}\!\left(
           \arg\max_c \cos(\tilde{q}_i,\boldsymbol{\mu}_c)
           \right)$
    \State $h_i \gets \beta h_i + (1-\beta)r_i$
    \State $g_i \gets
           \begin{cases}
               \operatorname{OneHot}(0), & x_i \in \mathcal{U}_t \text{ and } [f_v(z_i)]_0 \geq \tau_0,\\
               h_i, & x_i \in \mathcal{U}_t \text{ and } [f_v(z_i)]_0 < \tau_0
           \end{cases}$
    \State $s_i \gets
           \begin{cases}
               \operatorname{OneHot}(0), & x_i \in \mathcal{U}_t \text{ and } [f_v(z_i)]_0 \geq \tau_0,\\
               \operatorname{OneHot}(1), & x_i \in \mathcal{U}_t \text{ and } [f_v(z_i)]_1 \geq \tau_1,\\
               \text{Skip}, & x_i \in \mathcal{U}_t \text{ and neither condition holds}
           \end{cases}$
\EndFor
\State Compute $\mathcal{L}_{v}$ and $\mathcal{L}_{f}$ over the mini-batch
\State \Return $\mathcal{L}_{v}$ and $\mathcal{L}_{f}$
\end{algorithmic}
\end{algorithm}

\textbf{Refine Gate and Verifier (RAV).}
RAV is proposed to translate the discriminative representations learned by PAIR into reliable supervisory signals $s_i$. As shown in Algorithm~\ref{alg:rav}, RAV consists of two interacting components, namely the Verifier and the Refine Gate. The Verifier, implemented as an auxiliary classification head $f_v$, generates reliable supervisory signals $s_i$ for the frozen source classifier $\psi_s^{*}$. Meanwhile, the Refine Gate leverages historical pseudo-targets to construct refined targets $g_i$ for training the Verifier.

To generate reliable supervisory signals $s_i$ in Eq.~\eqref{eq:target_ce} for unlabeled buildings, the Verifier guides the update of historical pseudo-targets based on class prototypes. Each prototype $\boldsymbol{\mu}_c$ represents class $c$ in the embedding space. For each unlabeled building, the class whose current prototype has the higher cosine similarity to $\tilde{q}_i$ is encoded as the candidate pseudo-target $r_i$. For each sample $x_i$ in the current mini-batch, the prototype of its assigned class is then updated as follows:
\begin{equation}
\boldsymbol{\mu}_c \gets
    \alpha\boldsymbol{\mu}_c + (1-\alpha)\tilde{q}_i,
\label{eq:prototype_update}
\end{equation}
where $\alpha$ is the momentum coefficient. For simplicity, the normalization operator is omitted from Eq.~\eqref{eq:prototype_update}. Early in adaptation, both target-domain representations and class predictions are unstable, making candidate pseudo-targets susceptible to transient prediction noise. To mitigate this effect, the Refine Gate maintains a historical pseudo-target $h_i$ for each unlabeled sample. It initializes $h_i$ to $[0.5,\,0.5]^{\top}$ and aggregates class information across training iterations as:
\begin{equation}
h_i = \beta h_i + (1-\beta)r_i,
\label{eq:historical_target}
\end{equation}
where $\beta$ denotes the momentum coefficient.

However, when the target-domain class prior $\pi_t$ is unknown, historical aggregation alone cannot provide sufficiently reliable supervision. To avoid reliance on this prior, reliable negative supervisory signals are introduced during domain adaptation. Specifically, the Refine Gate uses the SAT strategy~\citep{wang2022freematch} to dynamically update the thresholds ${\tau}=[\tau_0,\tau_1]$ for the undamaged and damaged classes, eliminating the need to set them manually. For example, if the Verifier predicts that an unlabeled sample is undamaged and its confidence meets the class-adaptive threshold $\tau_0$, the Refine Gate sets the current pseudo-target $g_i$ to the negative-class label $[1,0]^{\top}$. For all other unlabeled buildings, the Refine Gate sets $g_i=h_i$. Specifically, this is formulated as follows:
\begin{equation}
g_i =
\begin{cases}
\operatorname{OneHot}(0),
& x_i \in \mathcal{U}_t \text{ and } [f_v(z_i)]_0 \geq \tau_0,\\
h_i,
& x_i \in \mathcal{U}_t \text{ and } [f_v(z_i)]_0 < \tau_0,
\end{cases}
\end{equation}
where $[f_v(z_i)]_0$ denotes the probability assigned by the Verifier $f_v$ to the negative class for an unlabeled target sample $x_i$. These high-confidence negative predictions apply only to the current iteration and do not overwrite the stored historical pseudo-targets. Consequently, incorrect negative assignments made early in adaptation can be corrected in later iterations as target-domain representations and class predictions improve. The Verifier's class predictions determine which prototype each unlabeled sample updates in Eq.~\eqref{eq:prototype_update}. In turn, the Refine Gate uses the prototype-based historical targets and the Verifier's negative-class probabilities to generate the refined targets $g_i$ for training the Verifier. The Verifier classification
loss is given by:

\begin{equation}
\mathcal{L}_{v}
=
\frac{1}{|\mathcal{P}_t|}
\sum_{x_i \in \mathcal{P}_t}
\operatorname{CE}\bigl(\mathbf{p}, f_v(z_i)\bigr)
+
\frac{1}{|\mathcal{U}_t|}
\sum_{x_i \in \mathcal{U}_t}
\operatorname{CE}\bigl(g_i, f_v(z_i)\bigr).
\label{eq:verifier_loss}
\end{equation}
The supervisory signal $s_i$ for the prediction $\psi_s^{*}(z_i)$ of the frozen source classifier is defined as follows:

\begin{equation}
\mathbf{s}_i =
\begin{cases}
\operatorname{OneHot}(0),
& x_i \in \mathcal{U}_t \text{ and } [f_v(z_i)]_0 \geq \tau_0,\\
\operatorname{OneHot}(1),
& x_i \in \mathcal{U}_t \text{ and } [f_v(z_i)]_1 \geq \tau_1,\\
\text{Skip},
& x_i \in \mathcal{U}_t \text{ and neither condition holds}.
\end{cases}
\label{eq:supervisory_signal}
\end{equation}
Unlabeled buildings that do not meet the threshold for their predicted
class are excluded from the $\mathcal{L}_f$. The classification objective in Eq.~\eqref{eq:target_ce} takes the following form over $\mathcal{P}_t$ and $\mathcal{U}_t$:

\begin{equation}
\mathcal{L}_{f}
=
\frac{1}{|\mathcal{P}_t|}
\sum_{x_i \in \mathcal{P}_t}
\operatorname{CE}\bigl(\mathbf{p}, \psi_s^{*}(z_i)\bigr)
+
\frac{1}{|\mathcal{U}_t|}
\sum_{x_i \in \mathcal{U}_t}
\operatorname{CE}\bigl(s_i, \psi_s^{*}(z_i)\bigr).
\label{eq:frozen_classifier_loss}
\end{equation}
The unlabeled term is omitted when no unlabeled sample in the mini-batch meets its class threshold.

\begin{algorithm}[!t]
\caption{Positive-unlabeled target-domain adaptation}
\label{alg:panda_adaptation}
\begin{algorithmic}[1]
\Require Target-domain datasets $\mathcal{P}_t$ and $\mathcal{U}_t$, feature extractor $\varphi$, online and target branches $\mathcal{F}$ and $\mathcal{F}'$, frozen source classifier $\psi_s^{*}$, verifier $f_v$, loss weights $w_r$ and $w_f$, and momentum coefficients $m$ and $\beta$

\Statex \textit{// Initialization}
\State Initialize $\{\boldsymbol{\mu}_c\}$ and set $h_i \gets [0.5, 0.5]^{\top}$ for all $x_i \in \mathcal{U}_t$
\For{each adaptation iteration}
    \State Sample $\mathcal{B}_P \subseteq \mathcal{P}_t$ and
           $\mathcal{B}_U \subseteq \mathcal{U}_t$

    \Statex \hspace{\algorithmicindent}\textit{// PAIR representation learning
            (Algorithm~\ref{alg:pair})}
    \State $\{z_i,\tilde{q}_i\},\mathcal{L}_{\mathrm{r}}
           \gets \Call{PAIR}{\mathcal{B}_P,\mathcal{B}_U}$

    \Statex \hspace{\algorithmicindent}\textit{// RAV supervision refinement
            (Algorithm~\ref{alg:rav})}
    \State $\mathcal{L}_v, \mathcal{L}_f
           \gets \Call{RAV}{\mathcal{B}_P,\mathcal{B}_U,
                            \{z_i,\tilde{q}_i\}}$

    \Statex \hspace{\algorithmicindent}\textit{// Target-domain model update}
    \State Update $\mathcal{F}$ and $f_v$ by minimizing
           $\mathcal{L}_{\mathrm{all}}$ in Eq.~\eqref{eq:overall_loss}
    \State Update the target branch by exponential moving average with momentum $m$
\EndFor

\State \Return $\psi_s^{*}\circ\varphi$
\end{algorithmic}
\end{algorithm}

The optimization objective $\mathcal{L}_{\mathrm{all}}$ for the target-domain adaptation stage of PANDA is formulated as:
\begin{equation}
\mathcal{L}_{\mathrm{all}}
=
\mathcal{L}_{v}
+
w_r\overline{\mathcal{L}}_{\mathrm{r}}
+
w_f\mathcal{L}_{f},
\label{eq:overall_loss}
\end{equation}
where $\overline{\mathcal{L}}_{\mathrm{r}}$ denotes the pairwise loss averaged over the sample pairs in the current mini-batch and can take one of three forms: $\mathcal{L}_{\mathrm{r}}^{\mathrm{s}}$, $\mathcal{L}_{\mathrm{r}}^{\mathrm{f}}$, or $\mathcal{L}_{\mathrm{r}}^{\mathrm{w}}$. As demonstrated in the subsequent experiments, $\mathcal{L}_{\mathrm{r}}^{\mathrm{w}}$ outperforms the other two alternatives due to its greater robustness to unreliable PU supervision. The coefficients $w_r$ and $w_f$ control the relative contributions of the corresponding loss terms. Algorithm~\ref{alg:panda_adaptation} summarizes the mini-batch adaptation procedure. The frozen source classifier $\psi_s^{*}$ is retained as the prediction head of PANDA during inference.

\section{Why target-domain discriminability matters for domain adaptation under distribution shifts}
\label{sec4}
Let $X_u$ and $Y_u$ denote the input and label in domain $u$, respectively,
and let $Z_t$ denote the target representation:
\begin{equation}
\label{eq:theory_setup}
Y_u \sim \mathrm{Ber}(\pi_u), \qquad
Y_u = \eta_u(X_u), \qquad
Z_t = \varphi(X_t), \qquad
u \in \{s,t\},
\end{equation}
where $\mathrm{Ber}$ denotes the Bernoulli distribution and $\pi_u = P(Y_u = 1)$ is the class prior for damaged buildings in domain $u$. In the target domain, $0 < \pi_t < 1$.
We assume that the input uniquely determines its label within each domain,
with $\eta_u$ denoting the corresponding labeling function.
Here, $X_u$ denotes the random input whose realizations are written as
$x$ in Eq.~\eqref{eq:mix}; the class-conditional distributions $P_u^d$
and $P_u^n$ correspond to conditioning on $Y_u=1$ and $Y_u=0$,
respectively.
The coupled distribution shifts comprise class-prior shift,
$\pi_s\neq\pi_t$, and class-conditional distribution shift,
$P_s(X\mid Y)\neq P_t(X\mid Y)$, corresponding to the
damage-proportion and damage-appearance shifts, respectively.

Here, $H(\cdot)$, $I(\cdot;\cdot)$, and $I(\cdot;\cdot\mid\cdot)$ denote entropy, mutual information, and conditional mutual information, respectively. Target-domain discriminability is characterized by $I(Y_t;Z_t)$~\citep{qiang2025transferability}, the target-label information retained in the representation; a larger value indicates that the representation contains more class information. Since the label entropy $H(Y_t)$ does not vary with the representation for a given target domain, the discriminability deficit is measured by the conditional entropy $H(Y_t\mid Z_t)=H(Y_t)-I(Y_t;Z_t)$, which represents the label uncertainty remaining given the representation. The transferability deficit is measured by $I(X_t;Z_t\mid X_s)$~\citep{qiang2025transferability}, the target-input information retained in the representation after conditioning on the source input; a smaller value indicates greater transferability under the specified joint distribution. The joint distribution of $X_s$ and $X_t$ is defined by the cross-domain coupling introduced below.
Let $K$ be a reference variable jointly distributed with $(X_t,Y_t)$
and representing source-related information.
By the chain rule for mutual information and $Y_t=\eta_t(X_t)$~\citep{cover2006elements},
\begin{equation}
\label{eq:information_chain}
I(Y_t;Z_t)
\le I(Y_t;K)+I(Y_t;Z_t\mid K)
\le I(Y_t;K)+I(X_t;Z_t\mid K).
\end{equation}
Subtracting both sides from $H(Y_t)$ and rearranging yields
\begin{equation}
\label{eq:information_tradeoff}
H(Y_t\mid Z_t)+I(X_t;Z_t\mid K)
\ge H(Y_t\mid K).
\end{equation}
For $K=X_s$, the two terms on the left are the discriminability
and transferability deficits, respectively, while the right-hand side
is the target-label uncertainty remaining given the source input.

\noindent\textbf{Class-prior shift.}
Consider a maximal coupling of $(Y_s,Y_t)$ such that
$P(Y_s\neq Y_t)=|\pi_s-\pi_t|$~\citep{levin2026markov}. 
Given the label pair, sample the inputs independently from their
respective class-conditional distributions, yielding the Markov chain
$X_s\to Y_s\to Y_t\to X_t$.
Together with $Y_s=\eta_s(X_s)$, this gives
$H(Y_t\mid X_s)=H(Y_t\mid Y_s)$.
When $\pi_s\neq\pi_t$, we have $0<P(Y_s\neq Y_t)<1$.
Thus, the nondegenerate binary variable $Y_t$ is neither $Y_s$ nor
its complement and cannot be a deterministic function of $Y_s$.
Consequently, $H(Y_t\mid Y_s)>0$, and
Eq.~\eqref{eq:information_tradeoff} implies
\begin{equation}
\label{eq:prior_shift_bound}
I(X_t;Z_t\mid X_s)=0
\quad\Longrightarrow\quad
H(Y_t\mid Z_t)\ge H(Y_t\mid Y_s)>0.
\end{equation}
Under this coupling, reducing uncertainty about target labels below the level achieved with the source input alone calls for the representation to retain additional target-label information.

\noindent\textbf{Class-conditional distribution shift.}
Let $\hat Y_t=\eta_s(X_t)$ denote the prediction obtained by
applying the source labeling function to the target input.
We consider class-conditional shifts under which the source rule
partially fails, i.e., $0<P_t(\hat Y_t\neq Y_t)<1$.
By the same argument, $H(Y_t\mid\hat Y_t)>0$.
Taking $K=\hat Y_t$ in Eq.~\eqref{eq:information_tradeoff} yields
\begin{equation}
\label{eq:conditional_shift_bound}
I(X_t;Z_t\mid\hat Y_t)=0
\quad\Longrightarrow\quad
H(Y_t\mid Z_t)\ge H(Y_t\mid\hat Y_t)>0.
\end{equation}
Here, $I(X_t;Z_t\mid\hat Y_t)=0$ means that the representation
retains only information explainable by the source prediction.
Since $Y_t=\eta_t(X_t)$, the conditional data processing inequality
gives $0\le I(Y_t;Z_t\mid\hat Y_t)
\le I(X_t;Z_t\mid\hat Y_t)=0$.
Thus, when the source rule partially fails, such a representation
likewise cannot retain all target-label information.

\noindent\textbf{Necessity of target-discriminative information.}
Let $\tilde Y_t=g(Z_t)$ be a prediction based on $Z_t$, with
target error $\varepsilon_t=P_t(\tilde Y_t\neq Y_t)$.
By the data processing inequality and binary Fano's inequality~\citep{cover2006elements},
\begin{equation}
\label{eq:target_fano}
h_b(\varepsilon_t)
\ge H(Y_t\mid\tilde Y_t)
\ge H(Y_t\mid Z_t),
\end{equation}
where $h_b$ is the binary entropy function.
Combining Eq.~\eqref{eq:target_fano} with
Eqs.~\eqref{eq:prior_shift_bound}
and~\eqref{eq:conditional_shift_bound}, we obtain
a target-error lower bound of $h_b^{-1}(H(Y_t\mid K))$
for representations retaining only information explainable
by the reference variable, under the respective assumptions.
Here, $h_b^{-1}$ denotes the inverse of $h_b$ on $[0,\tfrac12]$.
Achieving an error below this bound requires
$H(Y_t\mid Z_t)<H(Y_t\mid K)$.
By Eq.~\eqref{eq:information_chain},
\begin{equation}
\label{eq:target_information_necessity}
I(Y_t;Z_t\mid K)
\ge H(Y_t\mid K)-H(Y_t\mid Z_t)>0.
\end{equation}
These results reveal that source-related information alone may be insufficient to resolve target-label uncertainty under damage-proportion shift or damage-appearance shift. Under the respective assumptions, achieving a target error below the corresponding lower bound requires the representation to retain target-discriminative information beyond that provided by the reference variable. This motivates the use of weak supervision from the target domain, such as PU supervision, to enhance target-domain discriminability alongside the transfer of source knowledge.

\section{Experiments}
\label{sec5}

\subsection{Experimental settings}
\label{sec5.1}

\subsubsection{Implementation details}
\label{sec5.1.1}

Optimization was performed using stochastic gradient descent (SGD) with a momentum of 0.9 and a cosine annealing learning-rate schedule. The initial learning rate was set to 0.001 for all disaster events. The model was trained for 50 epochs with a batch size of 64 and 500 update steps per epoch. The hyperparameters $w_r$, $w_f$, $m$, and $\beta$ were set to 1, 1, 0.999, and 0.9, respectively, for all disaster events. All compared methods used ResNet-50 as the backbone. Each building instance was cropped using a square window centered on its bounding box, with an initial side length equal to the longer side of the box and an outward extension of 0.5 times this length in all directions to include surrounding context. Each crop was resized to $64 \times 64$ pixels and then to $224 \times 224$ pixels before being fed into the model.
\subsubsection{Evaluation metrics}
\label{sec5.1.2}

Overall Accuracy (OA) and $F_1$ score were used to evaluate building damage classification performance. OA reflects the overall proportion of correctly classified buildings, whereas $F_1$ provides a more informative assessment of damage detection under class imbalance, which is common in disaster scenarios. All experiments were independently repeated three times, and the mean and standard deviation of the results were reported.

\subsection{Baseline methods}
\label{sec5.2}

To evaluate PANDA, we used the Source-only model as a reference by applying it directly to the target-domain test set without adaptation, and compared PANDA with 12 methods across three categories. (1) source-free UDA methods. We selected SHOT~\citep{liang2020we}, TPDS~\citep{tang2024source}, DCPL~\citep{diamant2024confusing}, and PLUE~\citep{litrico2023guiding,litrico2026uncertainty}. (2) PU learning methods. We selected nnPU~\citep{kiryo2017positive}, ImbPU~\citep{su2021positive}, WSC~\citep{zhou2025weakly}, and NcPU~\citep{zhao2026noisy}. All but NcPU require the target-domain positive class prior.  (3) PU-based domain adaptation methods. We included PU-DA~\citep{sonntag2022positive}, a source-available domain adaptation method developed for intelligent transportation applications that requires the target-domain class prior. To evaluate the effect of directly combining source-domain pretraining with target-domain PU supervision, we also constructed three baselines initialized with the same pretrained source weights. Source + CE used cross-entropy loss, treating unlabeled samples as negative. Source + nnPU and Source + ImbPU used the nnPU and ImbPU objectives, respectively. For all baselines that require the prior, we used KM2~\citep{ramaswamy2016mixture} to estimate it.

\subsection{Comparison with baseline methods}
\label{sec5.3}

\input{Tab/main_results}

\begin{figure}[!t]
    \centering
    \includegraphics[width=1\linewidth]{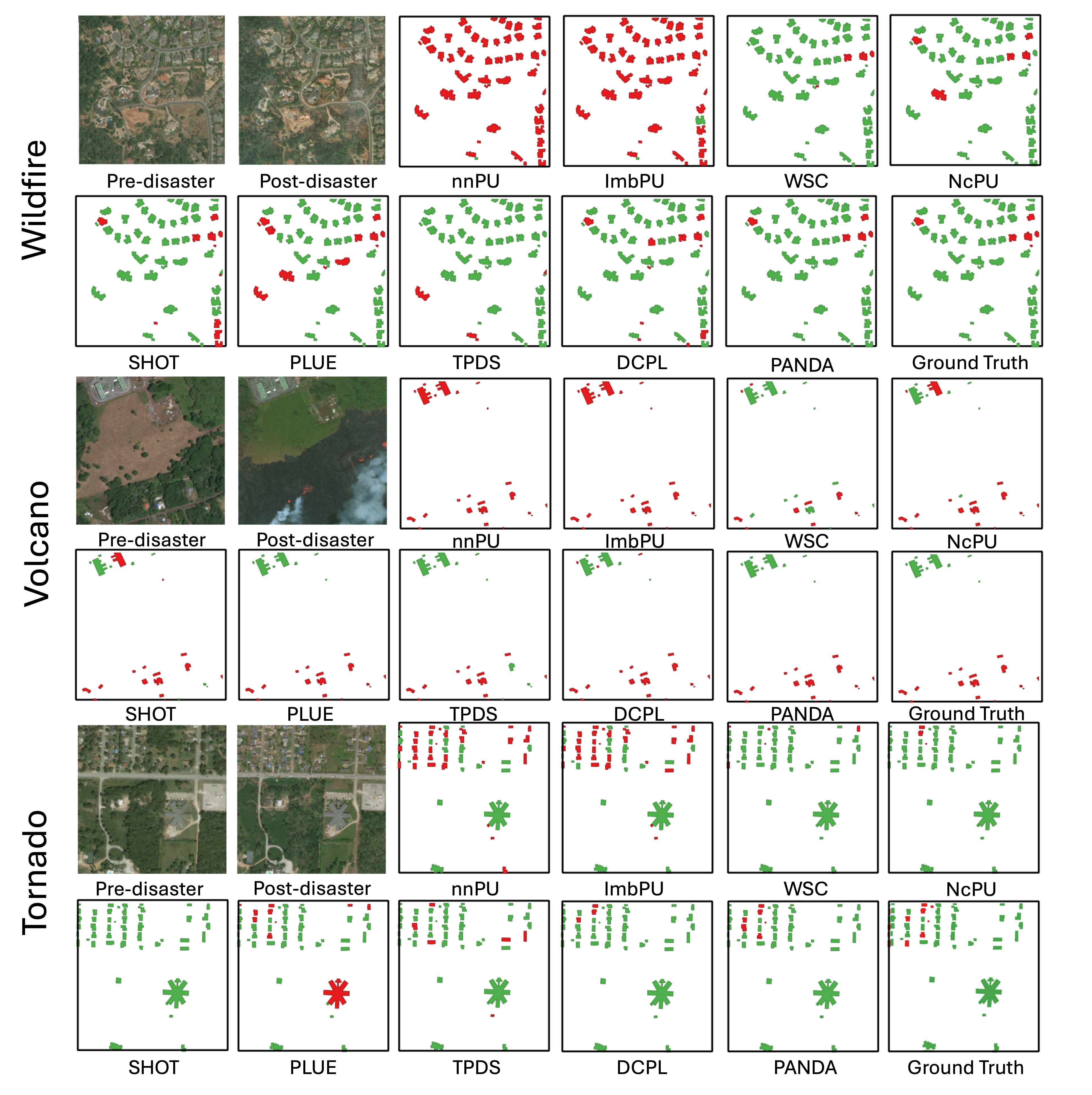}
    \caption{Visualization comparison of building damage mapping results on the target disaster events. Red and green indicate damaged and undamaged buildings, respectively.}
    \label{fig:baseline}
\end{figure}

Table~\ref{tab:main_results} presents the quantitative comparison between PANDA and the baseline methods on the three target-domain disaster events, while the corresponding visualization results are shown in Figure~\ref{fig:baseline}.

\textbf{Cross-event adaptation requires both transferability and target-domain discriminability.} As shown in Table~\ref{tab:main_results}, Source-only achieved $F_1$ scores of only 30.41\% to 41.29\% across the three target events, indicating that historical damage knowledge transferred poorly to new events without adaptation. Source-free UDA methods transfer historical damage knowledge using unlabeled target-domain data but lack target-domain class supervision. Their effectiveness may depend on the extent of the source-target distribution shift, and adaptation can even lead to negative transfer. For example, TPDS achieved an $F_1$ of 75.90\% on Volcano but only 24.30\% on Wildfire, below the Source-only result of 41.29\% on Wildfire. PU learning methods exploit target-domain supervision but lack historical damage knowledge. If damaged buildings in the unlabeled set $U$ exhibit damage patterns absent from the small set of confirmed positives $P$, the model may struggle to identify them. PU performance may therefore vary across events and with the choice of positive samples. ImbPU achieved $F_1$ scores of 84.74\% on Tornado and 46.93\% on Volcano. WSC also exhibited substantial run-to-run variability on Tornado due to instability late in training.

\textbf{Combining historical damage knowledge with PU supervision does not ensure reliable cross-event adaptation.} We evaluated PU-DA, which uses a two-stage adaptation procedure, and constructed three baselines that combine source-domain pretraining with target-domain PU supervision. We compared these methods to assess how effectively they integrate historical damage knowledge with target-domain discriminative information. As shown in Table~\ref{tab:main_results}, source-domain pretraining yielded only modest $F_1$ gains for nnPU and ImbPU on Tornado and Volcano. On Wildfire, Source + ImbPU exceeded ImbPU by only 0.29 percentage points in $F_1$, while Source + nnPU scored 8.19 percentage points lower than nnPU. PU-DA performance also varied across events, with $F_1$ scores of 66.94\% on Volcano and only 15.84\% on Wildfire.

\textbf{PANDA achieves the best overall performance across the three target events.} PANDA achieved $F_1$ scores of 91.68\%, 82.62\%, and 91.06\% on Tornado, Volcano, and Wildfire, respectively, and was the only method to exceed 80\% $F_1$ on all three events. PANDA outperformed all PU learning methods on each event, suggesting that historical damage knowledge improves target-domain damage recognition when few labeled positive samples are available. On Volcano, PANDA achieved an $F_1$ score of 82.62\%, outperforming the best source-free UDA baseline by 6.72 percentage points. This result suggests that target-domain PU supervision provides discriminative information that can guide the transfer of historical damage knowledge. PANDA also outperformed all PU-based domain adaptation baselines on each event, further indicating that the effectiveness of cross-event adaptation depends on how historical damage knowledge and target-domain PU supervision are combined.

\subsection{Ablation study and method analysis}
\label{sec5.4}
\subsubsection{Ablation analysis}
\label{sec5.4.1}
PANDA consists of two complementary components: (1) PAIR and (2) RAV. By default, PAIR uses the robust pairwise loss $\mathcal{L}_{\mathrm{r}}^{\mathrm{w}}$. To investigate the working mechanism of PANDA and assess the necessity and contribution of its individual components, ablation experiments were conducted on the Volcano and Tornado target events. All experiments followed the same training settings as the main experiments and were independently repeated with three random seeds.

\input{Tab/abla_main}

\input{Tab/abla_RAUPL}

\input{Tab/abla_RAV}

\textbf{Both PAIR and RAV contribute to reliable target-domain adaptation.}
To assess the contribution of each component, we removed PAIR and RAV separately and compared the resulting models with the complete PANDA. As shown in Table~\ref{tab:ablation_modules}, the complete model achieved the best overall performance. Removing RAV led to substantial performance degradation on both target events, indicating that representation alignment alone is insufficient for reliable target-domain adaptation. Removing PAIR also led to performance degradation, supporting the view that discriminative representation learning helps transform weak supervision into reliable supervision.

\textbf{PAIR enables more robust representation learning under cross-event distribution shifts.}
To evaluate this effect, we retained RAV and replaced $\mathcal{L}_{\mathrm{r}}^{\mathrm{w}}$ with either  $\mathcal{L}_{\mathrm{r}}^{\mathrm{s}}$ or $\mathcal{L}_{\mathrm{r}}^{\mathrm{f}}$. As shown in Table~\ref{tab:ablation_raupl}, $\mathcal{L}_{\mathrm{r}}^{\mathrm{w}}$ achieved the highest average performance on both target events while exhibiting relatively low variation across random seeds. Notably, $\mathcal{L}_{\mathrm{r}}^{\mathrm{f}}$ performed even worse than the instance-level $\mathcal{L}_{\mathrm{r}}^{\mathrm{s}}$, suggesting that cross-sample alignment without suppressing the influence of unreliable sample pairs may impair discriminative representation learning in the target domain, consistent with the analysis in Section~\ref{sec3.3.2}. In particular, $\mathcal{L}_{\mathrm{r}}^{\mathrm{w}}$ showed a more pronounced advantage on the more challenging Volcano event, further indicating that PAIR enables more robust use of cross-sample class relationships for target-domain representation learning.

\textbf{The interaction between the Verifier and the Refine Gate is critical
for RAV to obtain reliable target-domain pseudo-supervision.} To assess their respective contributions, we removed the Verifier and the Refine Gate separately and further constructed a variant that retains only historical aggregation in Eq.~(\ref{eq:historical_target}) without high-confidence negative correction, denoted as $h_i$. As shown in Table~\ref{tab:ablation_rav}, removing the Verifier led to substantial performance degradation on both target events, indicating that reliable target-domain class discrimination is fundamental to generating reliable pseudo-supervision. Removing the Refine Gate or using only $h_i$ also resulted in performance degradation, demonstrating that the Refine Gate uses high-confidence negative predictions to refine the Verifier's training targets $g_i$, improving target-domain damage classification. The complete RAV achieved the best performance on both target events.

\subsubsection{Integration with geospatial foundation models}
\label{sec5.4.2}
Advances in model architectures and large-scale pretraining enable geospatial foundation models (GFMs) to learn generalizable representations from extensive Earth observation data, providing transferable knowledge for downstream tasks~\citep{liu2024remoteclip,brown2025alphaearth,herzog2026olmoearth}. However, coupled distribution shifts across disaster events can still limit their generalization. To assess the performance of GFMs in cross-event building damage mapping and their potential integration with PANDA, we use RemoteCLIP~\citep{liu2024remoteclip} as a representative GFM and construct GFM-based baselines for this task. GFM is first fine-tuned with labeled source-domain data. We then compare three settings: GFM Source-only, which performs inference directly on the target domain without a transfer strategy; GFM + CE, which uses CE as the transfer strategy, treating unlabeled target-domain samples as negatives; and GFM + PANDA, which uses PANDA as the transfer strategy. CE and PANDA use the same target-domain positive and unlabeled samples, and all three settings follow the training and evaluation settings of the main experiments.

\input{Tab/GFM}

Given the class imbalance commonly observed in disaster scenarios, $F_1$ provides a more informative assessment of damage detection than OA. As shown in Table~\ref{tab:GFM}, GFM pretraining followed by source-domain fine-tuning is insufficient to fully address the coupled distribution shifts across disaster events. Target-domain adaptation using CE also remains limited. GFM + PANDA achieves the best overall performance among the GFM-based methods on both events. These results suggest that, although GFMs can learn generalizable representations, they still need an appropriate transfer strategy under cross-domain coupled distribution shifts. PANDA can serve as such a strategy, facilitating the transfer of both pretrained knowledge and historical damage knowledge. Future work could investigate whether parameter-efficient fine-tuning (PEFT) strategies tailored to GFMs can further improve cross-event adaptation.

\begin{figure}[!t]
    \centering
    \includegraphics[width=0.5\linewidth]{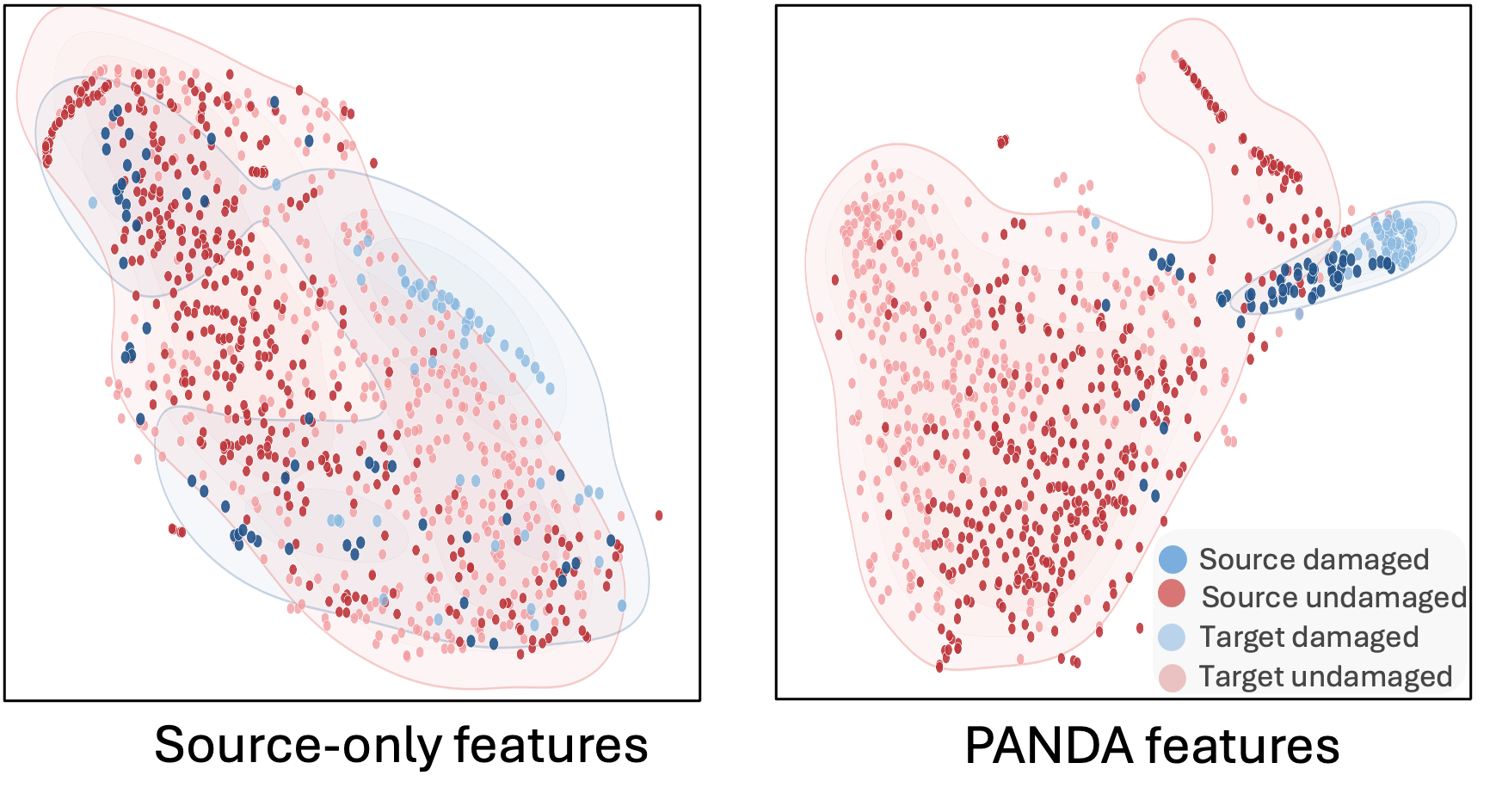}
    \caption{t-SNE visualization of feature representations learned by the Source-only model and PANDA for the source and target Volcano datasets. Colors indicate different classes, while dark and light circles denote source- and target-domain samples, respectively.}
    \label{fig:doubledomain}
\end{figure}

\begin{figure}
    \centering
    \includegraphics[width=1\linewidth]{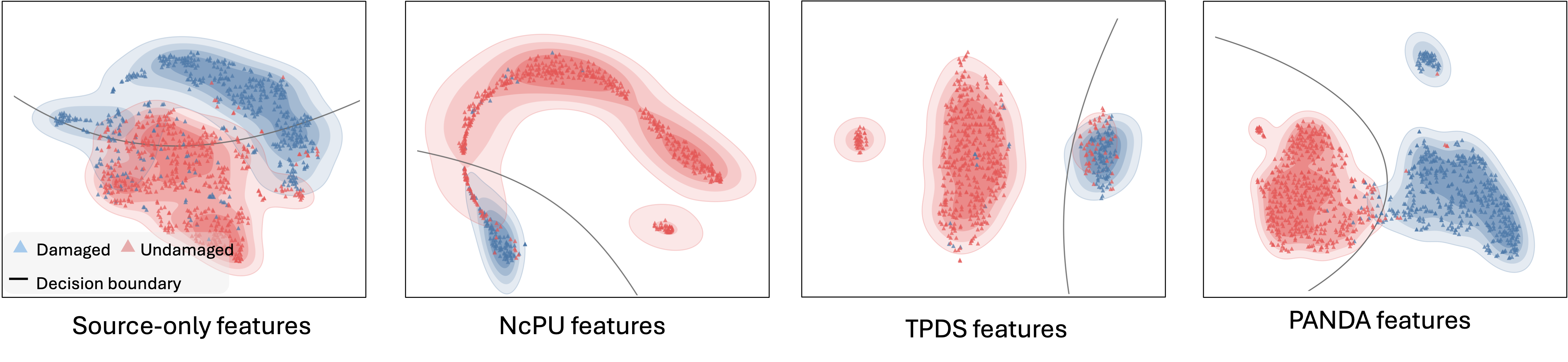}
    \caption{t-SNE visualizations of representations learned by baseline methods and PANDA on the target Volcano dataset.}
    \label{fig:features}
\end{figure}

\subsubsection{PANDA facilitates historical damage knowledge transfer while enhancing target-domain discriminability}
\label{sec5.4.3}
Figure~\ref{fig:doubledomain} shows that, compared with Source-only, PANDA achieves better alignment between source- and target-domain features within each class and clearer separation between classes. Figure~\ref{fig:features} further shows that baseline methods exhibit substantial class overlap in the target-domain feature space, whereas PANDA learns more clearly separable class representations. These patterns suggest that PANDA facilitates the transfer of historical damage knowledge while enhancing target-domain discriminability.

\begin{figure}[!t]
    \centering
    \includegraphics[width=1\linewidth]{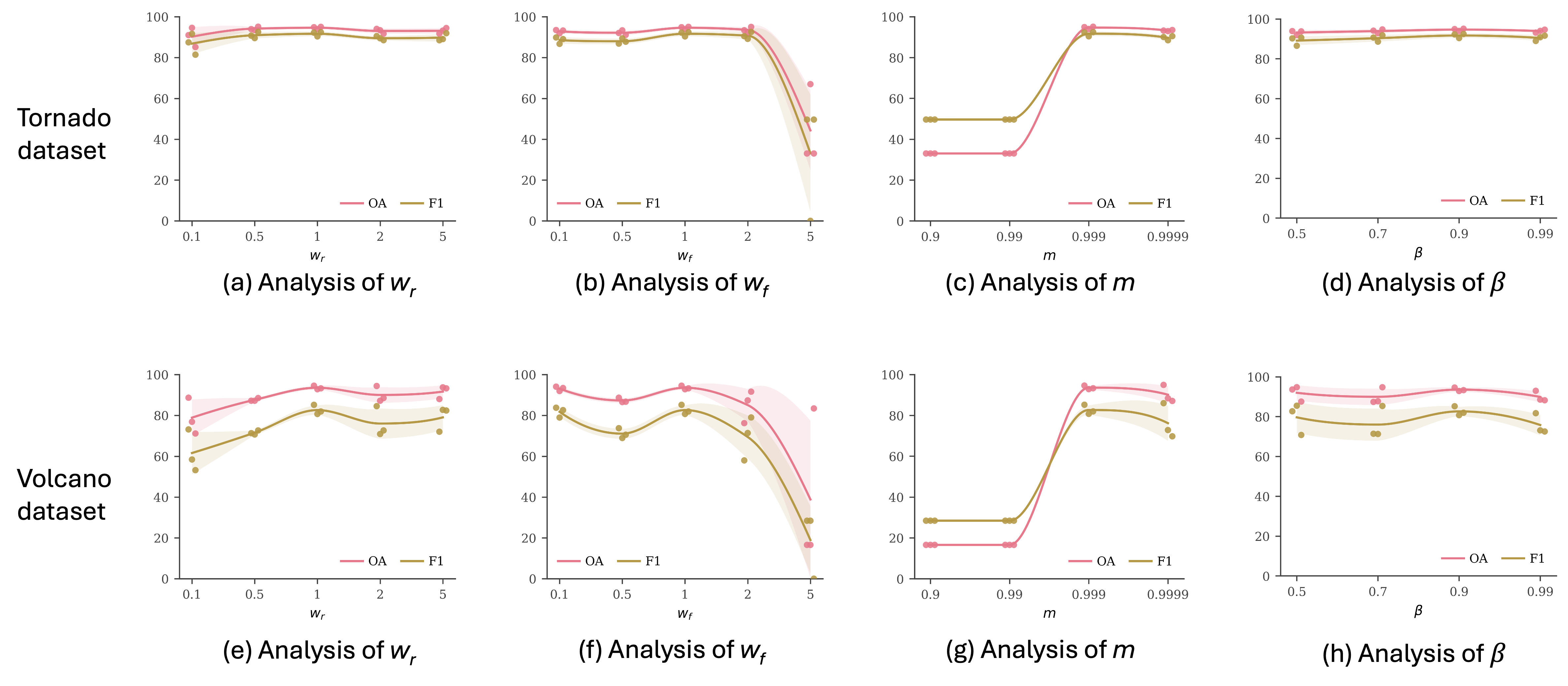}
    \caption{Hyperparameter analysis on the Tornado and Volcano  datasets.}
    \label{fig:Hy}
\end{figure}

\subsubsection{Analysis of hyperparameters}
\label{sec5.4.4}
As shown in Figure~\ref{fig:Hy}, PANDA remains stable over moderate ranges of $w_r$, $w_f$, and $\beta$, but is more sensitive to the momentum coefficient $m$. Performance generally improves as $w_r$ increases to 1, whereas an excessively large $w_f$ causes a sharp degradation because the frozen source classifier overly constrains target-domain adaptation. For $m$, small values lead to unstable performance, while $m=0.999$ yields the best results; in contrast, varying $\beta$ produces only limited performance changes, with the best overall results obtained at $\beta=0.9$. Overall, the same settings ($w_r=1$, $w_f=1$, $m=0.999$, and $\beta=0.9$) were used across all target events without event-specific hyperparameter tuning.

\subsubsection{Results with automatically extracted building footprints}
\label{sec5.4.5}
To further evaluate the applicability of PANDA when existing building footprint products are unavailable, we conducted experiments using automatically extracted building footprints. Specifically, we compared FootprintSAM and SAMPolyBuild in terms of building localization performance using Mask AP, Mask AP$_{75}$, and Segmentation Quality (SQ)~\citep{kirillov2019panoptic}, while building damage classification was assessed using OA and $F_1$ on target-domain test samples reconstructed from the automatically extracted footprints. Three target disaster events, i.e., Tornado, Volcano, and Wildfire, were considered.

\input{Tab/SAM_comprise}
\input{Tab/sam_mapping}

As shown in Table~\ref{tab:footprint_extraction}, FootprintSAM outperformed SAMPolyBuild across all three building localization metrics. Using the automatically extracted building footprints, PANDA achieved $F_1$ scores of 92.43\%, 61.04\%, and 92.10\% on Tornado, Volcano, and Wildfire, respectively (Table~\ref{tab:auto_footprint}). Performance on Volcano was lower than with reference footprints. These results demonstrate that, even when existing building footprint products are unavailable, PANDA can be effectively integrated with automatic building localization for building damage mapping, further validating its applicability in practical post-disaster response scenarios.

\subsection{Local-scale application and validation}
\label{sec5.5}
To evaluate the applicability of PANDA in real-world disaster response scenarios, we conducted local-scale validation on the Eaton Fire and Palisades Fire described in Section~\ref{sec2.2}.

\subsubsection{Application setup}
\label{sec5.5.1}

\textbf{Model adaptation and evaluation.}
For the local-scale case studies, the models followed the same settings as the main experiments. Section~\ref{sec5.5.2} further analyzes model performance under different numbers of positive samples. To further evaluate PANDA in practical disaster-response scenarios, we constructed a set of application-oriented baselines to compare different target-domain annotation costs and supervision settings. These included Source-only, Source + CE, and DCPL, which followed the same settings as in the main experiments. DCPL was selected for its consistently strong performance across the global benchmark. Two supervised learning baselines were also introduced. Under the matched annotation budget ($1\times$), $P/2$ positive and $P/2$ negative samples were used for supervised learning, resulting in the same total number of annotations as PANDA. Under the doubled annotation budget ($2\times$), $P$ positive and $P$ negative samples were used for supervised learning. These baselines were used to compare PANDA with fully supervised learning under the same and higher annotation costs. Evaluation and mapping covered all buildings in each study area.

\input{Tab/Local_results}

\subsubsection{Results}
\label{sec5.5.2}
\begin{figure}[!t]
    \centering
    \includegraphics[width=1\linewidth]{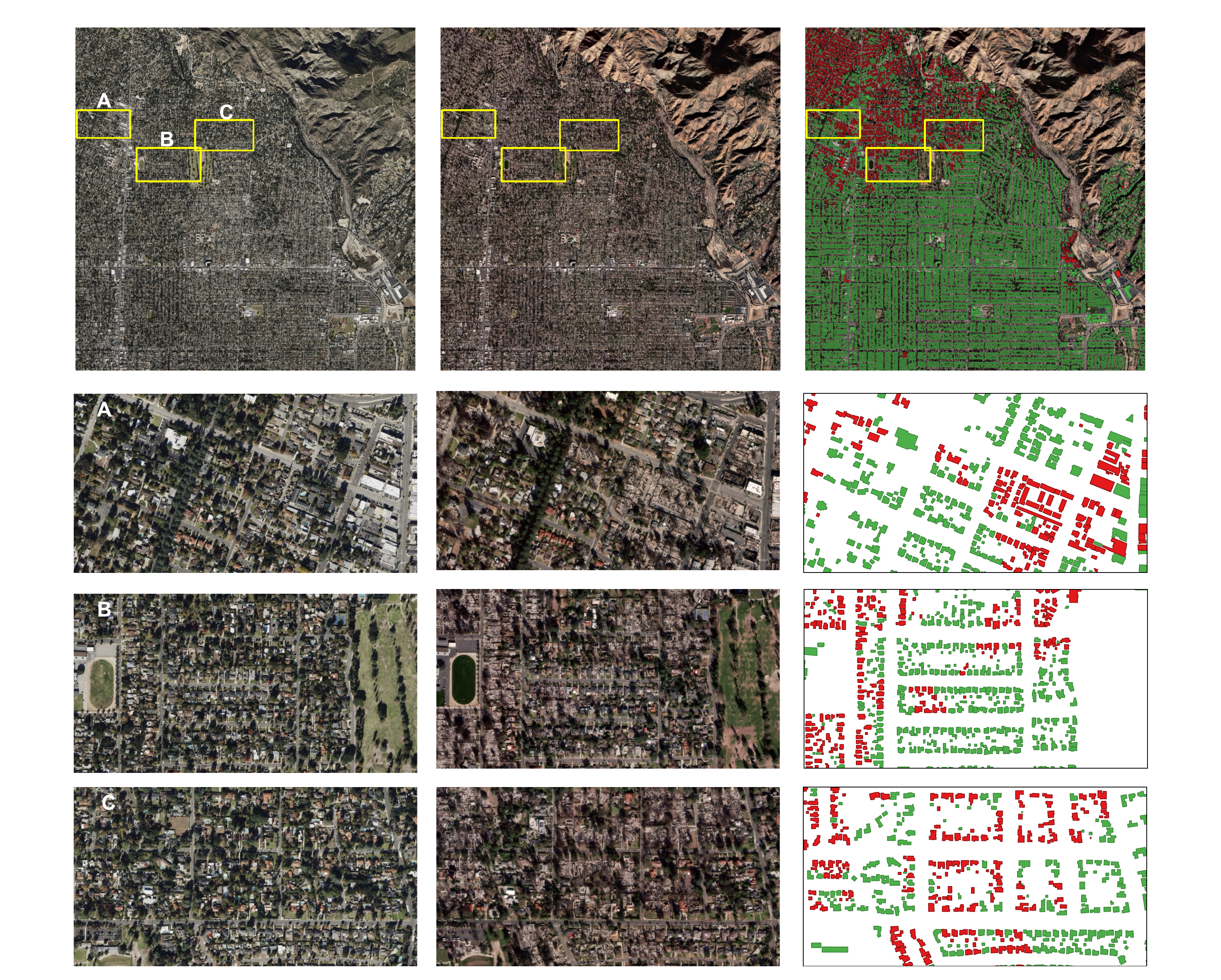}
    \caption{Visualization of building damage mapping results for the Eaton Fire in Los Angeles. Pre-disaster scenario, post-disaster scenario, and mapping results are shown from left to right, with three zones of interest (A–C) highlighted for detailed visualization. Red and green indicate damaged and undamaged buildings, respectively.}
    \label{fig:Eaton}
\end{figure}

\begin{figure}[!t]
    \centering
    \includegraphics[width=1\linewidth]{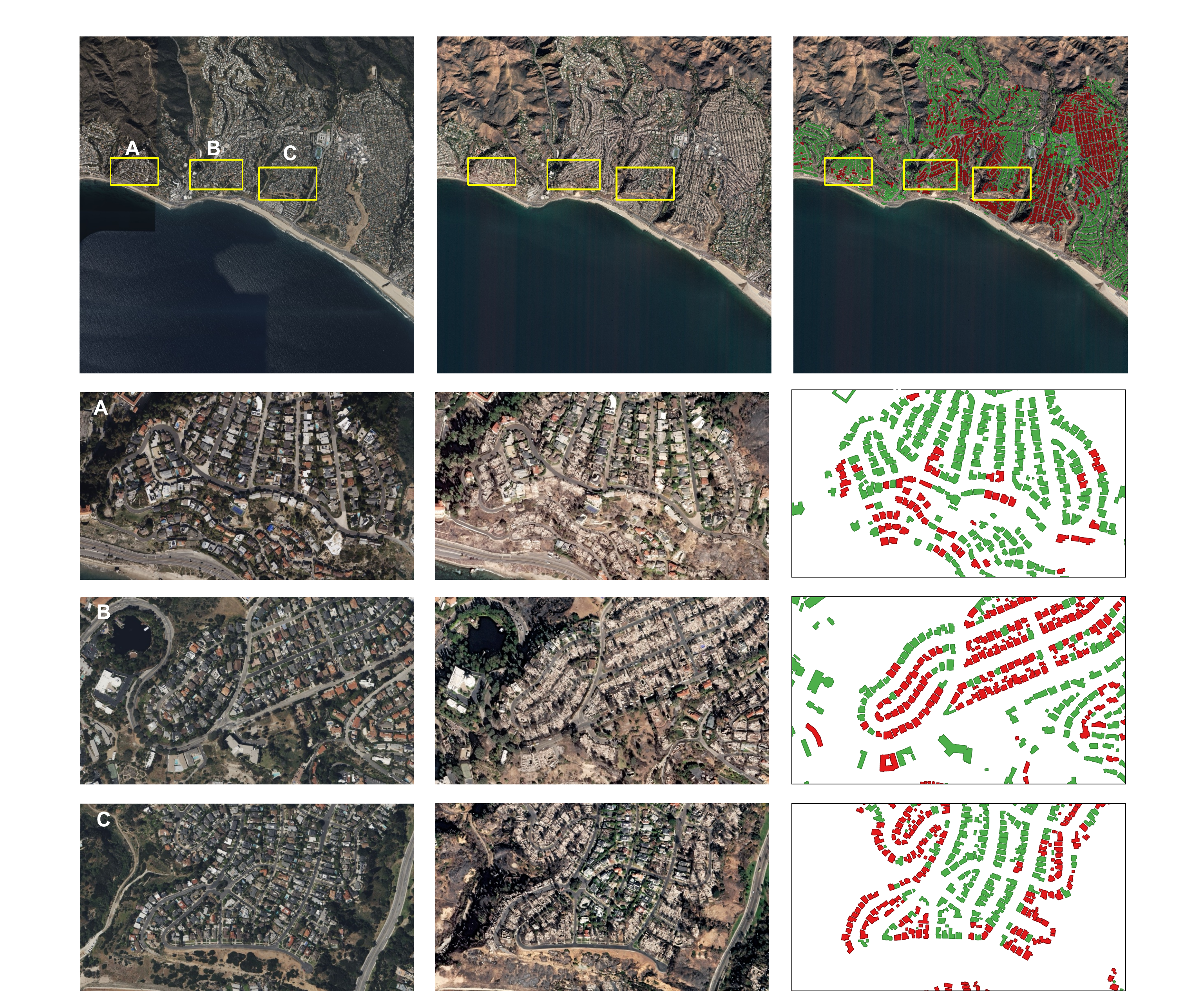}
    \caption{Visualization of building damage mapping results for the Palisades Fire in Los Angeles. Pre-disaster scenario, post-disaster scenario, and mapping results are shown from left to right, with three zones of interest (A–C) highlighted for detailed visualization. Red and green indicate damaged and undamaged buildings, respectively.}
    \label{fig:p}
\end{figure}

Table~\ref{tab:local_results} compares PANDA with the application-oriented baselines for building damage mapping in two real-world cases, the Eaton Fire (2025) and the Palisades Fire (2025). For the Eaton Fire
(Figure~\ref{fig:Eaton}), the study area contains 26,280 building instances, of which 3,752 are damaged. PANDA used only 50 damaged buildings as labeled target-domain positives, accounting for approximately 0.19\% of all building instances, while achieving higher performance than the supervised method with twice the annotation budget. For the Palisades Fire (Figure~\ref{fig:p}), the study area contains 7,799 building instances, including 5,073 damaged buildings. PANDA again used only 50 damaged buildings as labeled target-domain positives, representing approximately 0.64\% of all building instances, and achieved the best performance among all compared methods.

These results show that, under settings closer to practical disaster response, PANDA maintains strong cross-event adaptation capability while requiring only a very small amount of target-domain  annotation. This substantially reduces the annotation cost of target-domain adaptation and demonstrates the potential of PANDA for rapid building damage mapping in real-world disaster scenarios. 

\begin{figure}
    \centering
    \includegraphics[width=0.7\linewidth]{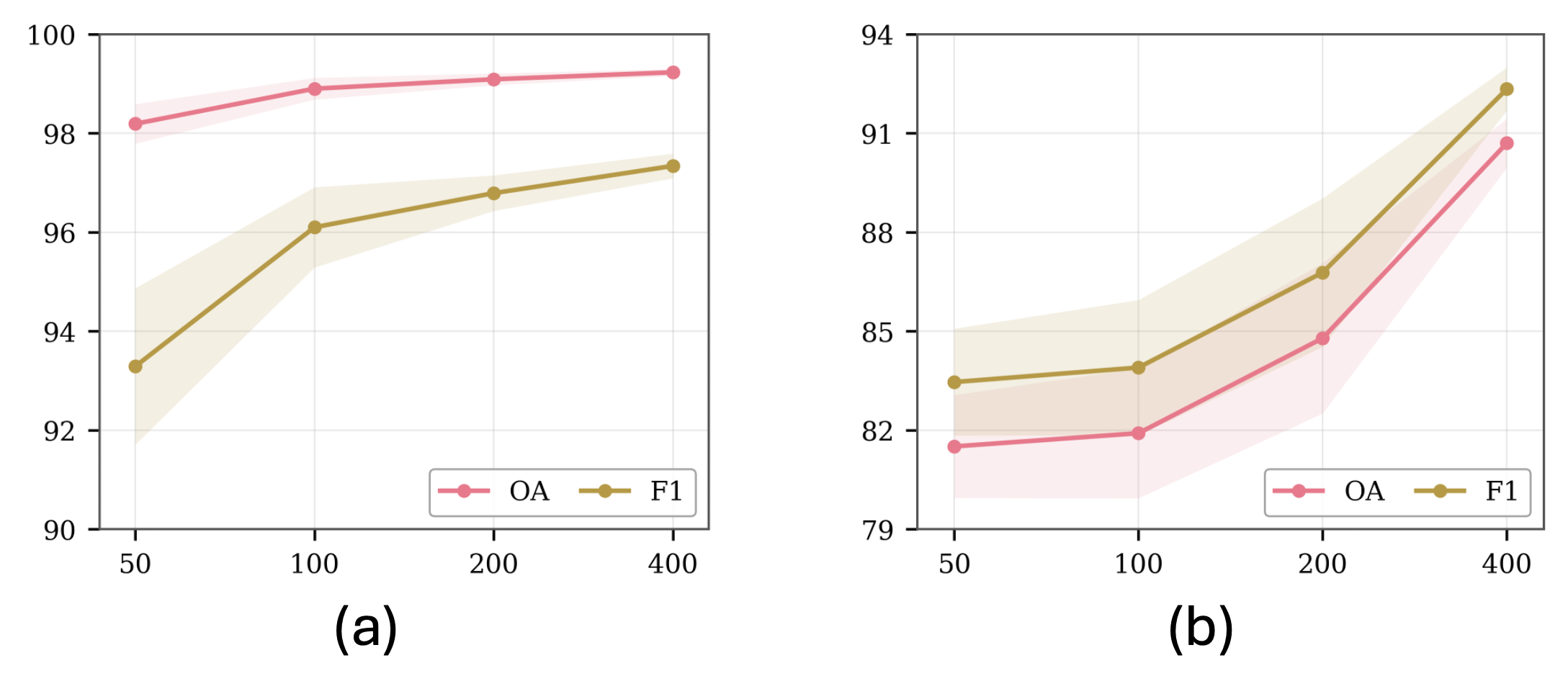}
    \caption{Sensitivity of PANDA to the number of positive samples \(P\) on (a) Eaton Fire and (b) Palisades Fire. Shaded regions denote the standard deviation across different runs.}
    \label{fig:pnum}
\end{figure}

We further analyzed the sensitivity of PANDA to the number of positive samples $P$. As shown in Figure~\ref{fig:pnum}, both OA and $F_1$ consistently improve as $P$ increases from 50 to 400 in the two study areas. For the Eaton Fire, PANDA already achieves strong performance with only 50
positive samples, and further increasing $P$ brings relatively limited gains. In contrast, the Palisades Fire is more sensitive to $P$, with the $F_1$ score increasing from 83.46\% to 92.34\%. These results indicate that a small number of positive samples is sufficient to support effective adaptation of PANDA, while additional positive samples can further improve performance in more challenging target scenarios.

For the Eaton Fire (Figure~\ref{fig:Eaton}), damaged buildings account for approximately 14.3\% of all building instances and are mainly distributed in spatially clustered areas, with clear transitions between severely affected communities and surrounding undamaged areas. In contrast, approximately 65.0\% of buildings are damaged in the Palisades Fire (Figure~\ref{fig:p}), where the affected areas exhibit a broader and more spatially continuous distribution. The enlarged local regions further show that PANDA can identify spatially contiguous damaged buildings in dense residential areas while preserving the distinction between adjacent undamaged buildings. The two cases exhibit markedly different damage proportions and spatial patterns, further demonstrating the adaptability of PANDA across diverse real-world disaster conditions.

\section{Conclusion}
\label{sec6}

This study presents a positive-unlabeled domain adaptation paradigm for cross-event building damage mapping across disaster events. Positive-unlabeled domain adaptation integrates domain adaptation with PU learning and requires only a small number of target-domain positive samples to enable rapid adaptation and mapping for new disaster events. Building on this paradigm, we propose PANDA, which combines PAIR for discriminative representation learning with RAV for target-domain supervision refinement. Together, these components promote cross-event transferability and target-domain discriminability for more reliable building damage mapping. Extensive experiments show that PANDA generally outperforms existing source-free UDA, PU learning and PU-based domain adaptation methods. The results also support PANDA’s potential for integration with GFMs. Two local-scale real-world disaster cases further demonstrate that accurate building damage maps can be produced using only a limited number of labeled damaged buildings, highlighting the potential of PANDA for rapid post-disaster response. In the future, PANDA could be extended to a broader range of weakly supervised remote sensing scenarios, providing a new approach to cross-domain remote sensing analysis under limited annotation.

\section*{Declaration of generative AI and AI-assisted technologies in the manuscript preparation process}

ChatGPT (OpenAI, GPT-6.1 Sol) was used to generate the panda illustrations and the damaged and undamaged building icons in Figure~2. The authors reviewed the illustrations and take full responsibility for the manuscript.

\section*{CRediT authorship contribution statement}
\textbf{Qirui Wang:} Writing - original draft, Visualization, Software, Methodology, Data curation, Conceptualization, Formal analysis, Investigation. \textbf{Hengwei Zhao:} Writing - review and editing, Data curation, Methodology, Conceptualization, Supervision. \textbf{Xiaoshu Yang:} Writing - review and editing, Data curation. \textbf{Jingtao Li:} Writing - review and editing. \textbf{Bo Li:} Writing - review and editing. \textbf{Sorin Popescu:} Writing - review and editing. \textbf{Rusty Feagin:} Writing - review and editing. \textbf{Wenzhe Jiao:} Writing - review and editing, Supervision, Resources, Funding acquisition.

\section*{Declaration of competing interest}
The authors declare that they have no known competing financial interests or personal relationships that could have appeared to influence the work reported in this paper.

\section*{Acknowledgments}
We acknowledge support from Agriculture and Food Research Initiative program (2024-68016-43058) of the United States Department of Agriculture National Institute of Food and Agriculture. Portions of this research were conducted with the advanced computing resources provided by Texas A\&M High Performance Research Computing.







\bibliographystyle{elsarticle-harv}
\bibliography{cas-refs}

\end{document}

%% file: Tab/main_results.tex
\begin{table*}[t]
\centering
\footnotesize
\setlength{\tabcolsep}{0.5pt}
\caption{Quantitative comparison of different methods on the target-domain disaster events.}
\label{tab:main_results}
\begin{tabular}{lcccccc}
\toprule
\multirow{2}{*}{Method}
& \multicolumn{2}{c}{Tornado}
& \multicolumn{2}{c}{Volcano}
& \multicolumn{2}{c}{Wildfire} \\
\cmidrule{2-3}
\cmidrule{4-5}
\cmidrule{6-7}
& OA & F1
& OA & F1
& OA & F1 \\
\midrule
Source-only
& $70.62^{\pm 5.03}$ & $38.57^{\pm 2.55}$
& $78.14^{\pm 4.37}$ & $30.41^{\pm 21.80}$
& $67.68^{\pm 10.00}$ & $41.29^{\pm 2.13}$ \\
\midrule
\multicolumn{7}{l}{\textit{source-free UDA methods}} \\
\midrule
SHOT
& $90.86^{\pm 0.53}$ & $87.37^{\pm 0.58}$
& $72.95^{\pm 2.31}$ & $54.01^{\pm 2.01}$
& $91.77^{\pm 2.45}$ & $84.72^{\pm 3.90}$ \\
TPDS
& $48.24^{\pm 16.85}$ & $43.04^{\pm 11.70}$
& $90.60^{\pm 1.84}$ & $75.90^{\pm 3.42}$
& $38.57^{\pm 10.44}$ & $24.30^{\pm 13.30}$ \\
DCPL
& $91.47^{\pm 0.45}$ & $88.15^{\pm 0.56}$
& $86.67^{\pm 1.29}$ & $67.20^{\pm 1.45}$
& $92.48^{\pm 0.75}$ & $85.44^{\pm 1.12}$ \\
PLUE
& $75.15^{\pm 2.00}$ & $70.99^{\pm 1.82}$
& $67.76^{\pm 2.89}$ & $50.40^{\pm 2.11}$
& $96.50^{\pm 0.48}$ & $92.85^{\pm 0.91}$ \\
\midrule
\multicolumn{7}{l}{\textit{PU learning methods}} \\
\midrule
nnPU
& $88.04^{\pm 0.42}$ & $84.32^{\pm 0.53}$
& $63.39^{\pm 2.33}$ & $47.45^{\pm 1.48}$
& $50.77^{\pm 2.57}$ & $47.62^{\pm 1.59}$ \\
ImbPU
& $88.45^{\pm 0.56}$ & $84.74^{\pm 0.61}$
& $62.47^{\pm 1.15}$ & $46.93^{\pm 0.83}$
& $58.79^{\pm 3.79}$ & $52.60^{\pm 2.44}$ \\
WSC
& $72.51^{\pm 8.10}$ & $25.44^{\pm 34.06}$
& $88.36^{\pm 0.82}$ & $50.95^{\pm 4.70}$
& $89.52^{\pm 2.40}$ & $74.68^{\pm 7.22}$ \\
NcPU
& $86.41^{\pm 4.38}$ & $73.92^{\pm 11.27}$
& $88.25^{\pm 1.89}$ & $71.47^{\pm 3.28}$
& $86.22^{\pm 0.55}$ & $57.62^{\pm 2.40}$ \\
\midrule
\multicolumn{7}{l}{\textit{PU-based domain adaptation methods}} \\
\midrule
Source + CE
& $67.05^{\pm 0.04}$ & $0.35^{\pm 0.24}$
& $88.31^{\pm 0.66}$ & $55.81^{\pm 4.64}$
& $76.99^{\pm 0.01}$ & $0.07^{\pm 0.12}$ \\
Source + nnPU
& $89.80^{\pm 0.19}$ & $86.29^{\pm 0.28}$
& $66.33^{\pm 1.26}$ & $49.52^{\pm 1.05}$
& $77.20^{\pm 0.62}$ & $39.43^{\pm 0.24}$ \\
Source + ImbPU
& $89.63^{\pm 0.62}$ & $86.09^{\pm 0.77}$
& $67.81^{\pm 0.58}$ & $50.71^{\pm 0.45}$
& $59.05^{\pm 1.36}$ & $52.89^{\pm 0.83}$ \\
PU-DA
& $73.46^{\pm 1.78}$ & $44.20^{\pm 10.70}$
& $90.11^{\pm 2.32}$ & $66.94^{\pm 9.19}$
& $78.80^{\pm 2.03}$ & $15.84^{\pm 14.91}$
\\
\midrule
PANDA
& $94.65^{\pm 0.60}$ & $91.68^{\pm 1.13}$
& $93.57^{\pm 0.91}$ & $82.62^{\pm 2.32}$
& $96.16^{\pm 0.16}$ & $91.06^{\pm 0.35}$ \\
\bottomrule
\end{tabular}
\end{table*}

%% file: Tab/abla_main.tex
\begin{table}[!t]
\centering
\footnotesize
\setlength{\tabcolsep}{6pt}
\caption{Ablation study of the PAIR and RAV modules.}
\label{tab:ablation_modules}
\begin{tabular}{cccccc}
\toprule
\multirow{2}{*}{PAIR} &
\multirow{2}{*}{RAV} &
\multicolumn{2}{c}{Volcano} &
\multicolumn{2}{c}{Tornado} \\
\cmidrule{3-4}
\cmidrule{5-6}
& & OA & F1 & OA & F1 \\
\midrule
\checkmark & --
& $66.07^{\pm 1.89}$ & $49.33^{\pm 1.52}$
& $69.64^{\pm 4.04}$ & $68.50^{\pm 2.79}$ \\
-- & \checkmark
& $88.74^{\pm 4.08}$ & $73.91^{\pm 7.30}$
& $95.26^{\pm 1.12}$ & $90.98^{\pm 1.34}$ \\
\checkmark & \checkmark
& $93.57^{\pm 0.91}$ & $82.62^{\pm 2.32}$
& $94.65^{\pm 0.60}$ & $91.68^{\pm 1.13}$ \\
\bottomrule
\end{tabular}
\end{table}

%% file: Tab/abla_RAUPL.tex
\begin{table}[!t]
\centering
\footnotesize
\setlength{\tabcolsep}{7pt}
\caption{Comparison of different representation alignment strategies in PAIR.}
\label{tab:ablation_raupl}
\begin{tabular}{ccccc}
\toprule
\multirow{2}{*}{PAIR} &
\multicolumn{2}{c}{Volcano} &
\multicolumn{2}{c}{Tornado} \\
\cmidrule{2-3}
\cmidrule{4-5} 
& OA & F1 & OA & F1 \\
\midrule
$\mathcal{L}_{\mathrm{r}}^{\mathrm{s}}$
& $91.48^{\pm 2.84}$ & $78.89^{\pm 5.35}$
& $94.63^{\pm 0.63}$ & $90.95^{\pm 1.07}$ \\
$\mathcal{L}_{\mathrm{r}}^{\mathrm{f}}$
& $89.46^{\pm 3.82}$ & $75.16^{\pm 7.00}$
& $93.93^{\pm 0.93}$ & $90.42^{\pm 1.72}$ \\
$\mathcal{L}_{\mathrm{r}}^{\mathrm{w}}$
& $93.57^{\pm 0.91}$ & $82.62^{\pm 2.32}$
& $94.65^{\pm 0.60}$ & $91.68^{\pm 1.13}$ \\
\bottomrule
\end{tabular}
\end{table}

%% file: Tab/abla_RAV.tex
\begin{table}[t]
\centering
\footnotesize
\setlength{\tabcolsep}{4pt}
\caption{Ablation study of the components in RAV.}
\label{tab:ablation_rav}
\begin{tabular}{cccccc}
\toprule
\multirow{2}{*}{Verifier} &
\multirow{2}{*}{Refine Gate} &
\multicolumn{2}{c}{Volcano} &
\multicolumn{2}{c}{Tornado} \\
\cmidrule{3-4}
\cmidrule{5-6}
& & OA & F1 & OA & F1 \\
\midrule
-- & \checkmark
& $52.13^{\pm 2.84}$ & $40.68^{\pm 1.40}$
& $40.03^{\pm 12.13}$ & $52.73^{\pm 5.35}$ \\
\checkmark & --
& $92.62^{\pm 1.23}$ & $78.76^{\pm 3.42}$
& $81.87^{\pm 2.55}$ & $62.53^{\pm 7.28}$ \\
\checkmark & $h_i$
& $62.24^{\pm 4.22}$ & $46.58^{\pm 2.69}$
& $90.98^{\pm 1.31}$ & $87.77^{\pm 1.45}$ \\
\checkmark & \checkmark
& $93.57^{\pm 0.91}$ & $82.62^{\pm 2.32}$
& $94.65^{\pm 0.60}$ & $91.68^{\pm 1.13}$ \\
\bottomrule
\end{tabular}
\end{table}

%% file: Tab/GFM.tex
\begin{table*}[t]
\centering
\footnotesize
\setlength{\tabcolsep}{0.5pt}
\caption{Performance of GFM-based methods for transferable building damage mapping.}
\label{tab:GFM}
\begin{tabular}{lcccc}
\toprule
\multirow{2}{*}{Method}
& \multicolumn{2}{c}{Volcano}
& \multicolumn{2}{c}{Tornado} \\
\cmidrule{2-3}
\cmidrule{4-5}
& OA & F1
& OA & F1 \\
\midrule
GFM (Source)
& $86.12^{\pm 0.50}$ & $54.44^{\pm 2.45}$
& $88.71^{\pm 1.05}$ & $79.94^{\pm 2.29}$ \\
GFM (CE)
& $88.42^{\pm 1.23}$ & $55.79^{\pm 7.47}$
& $67.04^{\pm 0.05}$ & $0.28^{\pm 0.32}$ \\
GFM (PANDA)
& $92.46^{\pm 3.56}$ & $81.08^{\pm 6.83}$
& $92.55^{\pm 0.92}$ & $89.37^{\pm 0.98}$ \\
\bottomrule
\end{tabular}
\end{table*}

%% file: Tab/SAM_comprise.tex
\begin{table}[t]
\centering
\footnotesize
\setlength{\tabcolsep}{6pt}
\caption{Comparison of building footprint extraction performance on the pooled test samples from all target-domain datasets.}
\label{tab:footprint_extraction}
\begin{tabular}{lccc}
\toprule
Method & Mask AP & Mask AP$_{75}$ & SQ \\
\midrule
SAMPolyBuild & 25.79 & 21.30 & 75.09 \\
FootprintSAM & 28.03 & 28.76 & 77.23 \\
\bottomrule
\end{tabular}
\end{table}

%% file: Tab/sam_mapping.tex
\begin{table}[t]
\centering
\footnotesize
\caption{Performance of PANDA using automatically extracted building footprints.}
\label{tab:auto_footprint}
\begin{tabularx}{0.92\linewidth}{
@{}
>{\centering\arraybackslash}p{0.20\linewidth}
*{4}{>{\centering\arraybackslash}X}
@{}
}
\toprule
\multirow{2}{*}{Disaster Event} &
\multicolumn{2}{c}{Building Localization} &
\multicolumn{2}{c}{Building Damage Classification} \\
\cmidrule{2-3}
\cmidrule{4-5}
& Mask AP & Mask AP$_{75}$ & OA & F1 \\
\midrule
Tornado  & 38.88 & 39.40 & $95.03^{\pm 0.21}$ & $92.43^{\pm 0.40}$ \\
Volcano  & 33.75 & 31.97 & $84.80^{\pm 0.52}$ & $61.04^{\pm 1.49}$ \\
Wildfire & 32.55 & 28.26 & $96.38^{\pm 0.31}$ & $92.10^{\pm 0.90}$ \\
\bottomrule
\end{tabularx}
\end{table}

%% file: Tab/Local_results.tex
\begin{table}[!t]
\centering
\footnotesize
\setlength{\tabcolsep}{6pt}
\renewcommand{\arraystretch}{1.15}
\caption{Quantitative comparison on the local wildfire events.}
\label{tab:local_results}
\begin{threeparttable}
\begin{tabular}{lcccc}
\toprule
\multirow{2}{*}{Method} &
\multicolumn{2}{c}{Eaton Fire} &
\multicolumn{2}{c}{Palisades Fire} \\
\cmidrule{2-3}
\cmidrule{4-5}
& OA & F1 & OA & F1 \\
\midrule
Source
& $85.72^{\pm 0.01}$ & $0.47^{\pm 0.78}$
& $34.98^{\pm 0.05}$ & $0.09^{\pm 0.16}$ \\

Source + CE
& $86.85^{\pm 0.53}$ & $14.87^{\pm 6.57}$
& $36.08^{\pm 0.33}$ & $3.43^{\pm 1.00}$ \\

DCPL
& $83.05^{\pm 2.25}$ & $62.54^{\pm 3.12}$
& $70.49^{\pm 15.63}$ & $71.83^{\pm 16.58}$ \\

Supervised ($1\times$)
& $90.84^{\pm 0.64}$ & $52.92^{\pm 5.06}$
& $51.95^{\pm 4.96}$ & $41.11^{\pm 9.32}$ \\

Supervised ($2\times$)
& $96.06^{\pm 0.86}$ & $84.14^{\pm 4.16}$
& $79.39^{\pm 2.92}$ & $81.23^{\pm 3.15}$ \\

\midrule
PANDA
& $98.19^{\pm 0.40}$ & $93.29^{\pm 1.58}$
& $81.51^{\pm 1.56}$ & $83.46^{\pm 1.62}$ \\
\bottomrule
\end{tabular}
\end{threeparttable}
\end{table}